\documentclass[11pt, preprint]{elsarticle}
\usepackage[a4paper, top=3cm, bottom=3cm, left=2cm, right=2cm]{geometry}
\usepackage[utf8]{inputenc}
\usepackage[english]{babel}
\usepackage{hyperref}
\usepackage{soul}
\usepackage{amsmath, amssymb}
\usepackage{mathtools}
\usepackage{tabularx}
\usepackage{subcaption}
\usepackage{standalone}
\usepackage{adjustbox}
\usepackage{makecell}
\usepackage{natbib}
\DeclarePairedDelimiterX{\infdivx}[2]{(}{)}{%
  #1\;\delimsize\|\;#2%
}
\newcommand{\infdiv}{KL\infdivx}

\newcommand*{\prob}{\mathsf{P}}
\DeclareMathOperator*{\argminA}{arg\,min}

\newcommand{\E}{{\rm I\kern-.3em E}}

\date{\today}
\begin{document}
\begin{frontmatter}

\title{A Benchmark on Uncertainty Quantification for Deep Learning Prognostics}

\author[onera]{Luis Basora}
\author[onera]{Arthur Viens}
\author[zhaw]{Manuel Arias Chao}
\author[onera]{Xavier Olive}

\address[onera]{ONERA DTIS, Université de Toulouse, France}
\address[zhaw]{Zurich University of Applied Sciences, School of Engineering, Winterthur, Switzerland}

\begin{abstract}
Reliable uncertainty quantification on RUL prediction is crucial for informative decision-making in predictive maintenance. In this context, we assess some of the latest developments in the field of uncertainty quantification for prognostics deep learning. This includes the state-of-the-art variational inference algorithms for Bayesian neural networks (BNN) as well as popular alternatives such as Monte Carlo Dropout (MCD), deep ensembles (DE) and heteroscedastic neural networks (HNN). All the inference techniques share the same inception deep learning architecture as a functional model. We performed hyperparameter search to optimize the main variational and learning parameters of the algorithms. The performance of the methods is evaluated on a subset of the large NASA N-CMAPSS dataset for aircraft engines. The assessment includes RUL prediction accuracy, the quality of predictive uncertainty, and the possibility to break down the total predictive uncertainty into its aleatoric and epistemic parts. The results show no method clearly outperforms the others in all the situations. Although all methods are close in terms of accuracy, we find differences in the way they estimate uncertainty. Thus, DE and MCD generally provide more conservative predictive uncertainty than BNN. Surprisingly, HNN can achieve strong results without the added training complexity and extra parameters of the BNN. For tasks like active learning where a separation of epistemic and aleatoric uncertainty is required, radial BNN and MCD seem the best options.
\end{abstract}

\begin{keyword}
Predictive Maintenance; Remaining Useful Life; Uncertainty Quantification; Bayesian Neural Network; Variational Inference; Monte Carlo Dropout; Deep Ensembles; Heteroscedastic Neural Network
\end{keyword}

\end{frontmatter}


\section{Introduction}
Unlike traditional corrective and scheduled maintenance, condition-based and predictive maintenance use models to identify from sensor data when a system performance reaches an unsatisfactory level. In a predictive maintenance (PM) strategy, maintenance decisions are based on an estimation of the current state of the system (diagnostics) and predictions of the failure time (prognostics) \cite{LEE2014314}. Hence, from a technical standpoint, finding cost-optimal maintenance intervention strategies ultimately depends on reliable prognostic algorithms estimating the system's remaining useful lifetime (RUL).
 
The application of machine learning and deep learning techniques for RUL estimation is an active area of research. The lack of uncertainty quantification (UQ) in most of the current state-of-the-art deep learning methods is a major drawback for their adoption in critical decision systems. This is the case in prognostics, where informative decision-making for PM depends not only on good prediction accuracy but also on the quality of predictive uncertainty. This is especially true because of the black-box nature of deep learning models, whose predictions are often overconfident and difficult, if not impossible, to interpret.

In prognostics, we need to address the two main categories of uncertainties introduced by \citet{der2009aleatory} and \citet{kendall2017uncertainties}: epistemic and aleatoric uncertainty. Epistemic or model uncertainty accounts for uncertainty in the model and can be reduced with additional data, improved methods, or a better understanding of the system being modeled. In practice, however, it is hard to gather enough data to cover for the diversity of operating situations leading to the different failure modes. On the other hand, aleatoric uncertainty cannot be eliminated but only characterized, as it is the result of the inherent variability introduced by a combination of different sources of noise, e.g., in sensor readings, manufacturing, and operational processes. Aleatoric uncertainty can be further classified as homoscedastic or heteroscedastic depending on whether it is constant, or evolves over the lifetime of the system like in prognostics. From a pragmatic point of view, it is useful to quantify epistemic and aleatoric uncertainty separately. For instance, epistemic uncertainty may guide the gathering of new data in active learning \cite{gal2017deep, ZHU2022108758} and support cost-sensitive decision-making  \cite{amodei2016concrete}.

Neural networks have become prominent at mean estimation in regression tasks, but there is evidence as well that they tend to make overconfident predictions \cite{guo2017calibration} with potentially harmful effects \cite{amodei2016concrete}. Recent research \cite{skafte2019reliable,seitzer2022pitfalls} has shown that predictive variance estimation is, in fact, a fundamentally different and more challenging task for neural networks, especially when a single network needs to estimate both mean and variance. In a typical Gaussian heteroscedastic approach, the input-dependent mean and variance are estimated by a neural network whose parameters are learned via maximum likelihood estimation (MLE). However, the minimization of the negative likelihood (NLL) by MLE produces overconfident variance estimates, and suboptimal mean predictions \cite{seitzer2022pitfalls}. In the paper, we refer to this type of neural network as a heteroscedastic neural network (HNN).

Bayesian Deep Learning (BDL) has been the focal point in the deep learning community for research on UQ methods. In particular, Bayesian neural networks (BNN) have drawn considerable attention in recent years. The main difference with a standard deterministic neural network is that BNN weights are random variables. The epistemic uncertainty is introduced with a prior distribution over the weight space and a posterior distribution inferred from data. Heteroscedastic uncertainty is usually modeled with a parametric likelihood distribution, whose parameters are the BNN outputs.

The posterior predictive distribution is obtained through weight marginalization, but its exact computation is intractable for neural networks. Thus, BDL research has focused on inference algorithms to approximate the predictive distributions, especially stochastic variational inference (VI) \cite{blundell2015weight}. Alternatively, Monte Carlo Dropout (MCD) \cite{gal_dropout_2016} or deep ensembles (DE) \cite{lakshminarayanan_simple_2017} have become popular for practical reasons in spite of not being strictly Bayesian. 

However, the application of these UQ techniques to deep learning RUL prediction has not been thoroughly studied. Comparative studies in the area are limited with regard to the number of methods tested or the aspects of the performance assessed (see Section \ref{sec:literature} for further details). Indeed, most of the benchmarks in the field of UQ for deep learning focus on classification (e.g., \citet{ovadia2019can}), whereas regression is only covered with the small benchmark UCI datasets and basic neural networks. Therefore, the research question we address in this paper is the following: In the context of UQ for deep learning prognostics, how do some of the most popular UQ techniques (HNN, BNN, MCD, and DE) compare with each other in terms of RUL prediction accuracy and quality of predictive uncertainty?

\textbf{Our contribution}. We provide a benchmark of UQ methods for deep learning prognostics, covering some recent advances on VI for BNN (including Radial BNN) as well as popular alternatives like HNN, MCD, and DE. All these techniques share the same inception deep learning architecture \cite{szegedy2015going, devol2021inception} as a functional model, and we optimize their most important hyperparameters with the Tree-structured Parzen Estimator (TPE) algorithm \cite{bergstra2011algorithms}. We assess the performance of the resulting models by using proper scoring rules \cite{gneiting2007strictly} on the recent NASA N-CMAPSS dataset \cite{arias2021aircraft} for turbofan engines, which is specifically designed for deep learning RUL estimation. The quality of predictive uncertainty is measured in terms of calibration and sharpness \cite{dawid1982well, degroot1983comparison} in different settings, including out-of-distribution (OOD) data. The benchmark is designed to provide answers to the following questions:

\begin{itemize}
\item How different are the considered methods in terms of RUL accuracy?
\item How trustworthy are the uncertainty estimates of these methods?
\item How does method accuracy evolve by progressively removing the samples a model is the least confident about?
\item Are the uncertainty estimates robust against challenging failure modes or with OOD settings?
\item Is their performance uniform over the different categories of data, e.g., failure modes, aircraft engine types? 
\item How does method performance evolve over the system lifetime?
\item How the break-down into epistemic and aleatoric uncertainty compares among the methods? 
\item What about other performance criteria, like training convergence time in training? 
\end{itemize}

In addition, for reproducibility and future research purposes, the code for our benchmark framework is released in open-source at \url{https://github.com/lbasora/bayesrul}. The design of the framework is extensible to include new methods and datasets.  

The paper is organized as follows. Section \ref{sec:literature} reviews the literature to help define the scope of the benchmark in terms of the methods and deep learning architectures to evaluate. Section \ref{sec:framework} describes the UQ methods, model architectures, metrics, and uncertainty decomposition techniques that will be used in the benchmark. Section \ref{sec:results} presents and discusses the results. Finally, a summary of the current and future work is given in Section \ref{sec:conclusion} 

\section{Literature review} \label{sec:literature}
The purpose of this literature review is two-fold. First, we review the state-of-the-art UQ methods for deep learning prognostics to consider in the benchmark. Secondly, we identify other existing benchmarks on the same domain to help focus our contribution and avoid duplication. 

\subsection{Deep learning techniques in prognostics}
The application of deep learning to prognostics steadily increased over the last few years because of its performance improvement over classical techniques for complex datasets \cite{fink2020potential, biggio2020prognostics}. Deep learning can deal with large volumes of high-dimensional data and automatically extract features from raw sensor data through representation learning. Several deep learning architectures have been proposed, especially deep convolutional neural networks \cite{li2018remaining, ren2018prediction} and recurrent neural networks (RNN) \cite{wu2018remaining, chen2019gated, wu2020data}. Some deep learning architectures have specifically been tested with C-MAPSS, such as the linear D3 \cite{benker2021utilizing} and convolutional C2P2 architecture \cite{sateesh2016deep, benker2021utilizing}. In the 2021 challenge of the annual PHM conference, recent deep learning techniques have been applied to N-CMAPSS, such as convolutional models to deal with variable-length sequences \cite{lovberg2021remaining} or inception architectures \cite{szegedy2015going, devol2021inception}. Finally, we can also cite some recent and sophisticated deep learning methods for RUL estimation, such as one combining recurrent neural networks with domain adversarial neural networks \cite{ganin2016domain, da2020remaining}, or another one based on capsule neural networks \cite{sabour2017dynamic, ruiz2020novel}.

\subsection{UQ for deep learning}
In spite of the advances in deep learning, research focuses mostly on improving point-estimate prediction accuracy rather than uncertainty estimation. However, there exists in the literature a number of approaches to measure uncertainty in neural networks \cite{gawlikowski2021survey,abdar2021review}.

Aleatoric UQ in neural networks model originates in 1994 with the contributions of \citet{nix1994estimating} and \citet{bishop1994mixture}, where predictive mean and variance are modeled as two separate neural networks, often as multi-layer perceptrons (MLP). Some difficulties and solutions on estimating the heteroscedastic variance with neural networks are reported in recent work \cite{skafte2019reliable, seitzer2022pitfalls}.

Regarding early work in BNN, we can refer to the contribution \citet{mackay1992practical} as early as 1992. BNN can capture epistemic uncertainty by putting a prior distribution over the neural network weights and then inferring an approximate posterior distribution from data. The posterior predictive distribution is obtained by marginalization, whose exact calculation is intractable. Thus, most of the research effort focused on inference methods to approximate such distribution such as Hamiltonian Monte Carlo (HMC) \cite{neal2011mcmc}, Laplace approximation \cite{ritter2018scalable}, expectation-propagation \cite{hernandez2015probabilistic} and stochastic variational inference (VI) \cite{graves2011practical, blundell2015weight}.

Recently, Variational Inference (VI) methods have drawn a lot of attention, especially mean-field variational inference (MFVI). MFVI can be implemented via Bayes by Backprop \cite{blundell2015weight} along with techniques to reduce gradient variance such as the local reparametrization trick (LRT) \cite{kingma2015variational} or Flipout \cite{wen2018flipout}. More recently, Radial BNN \cite{farquhar2020radial} is introduced as an inference technique solving the sampling problem of MFVI (also referred to as the `soap-bubble'), which according to the authors, results in better scalability. Low-rank \cite{tomczak2020efficient} is another recent scalable VI algorithm that learns Gaussian variational posteriors with non-diagonal covariance matrices by extending the LRT. 

VI methods scale better compared to Markov Chain Monte Carlo (MCMC), which is the traditional alternative to estimate the posterior predictive distribution. Even though there are more efficient MCMC variants in high dimensional space like Hamiltonian Monte Carlo \cite{neal2011mcmc}, VI is still a more widely used technique in Bayesian deep learning.

State-of-the-art approximate Bayesian techniques such as MCD \cite{gal_dropout_2016} and DE \cite{lakshminarayanan_simple_2017} are often presented as more practical alternatives due to their simplicity, scalability, and computational efficiency \cite{kefalasend2022}. However, there is a theoretical debate \cite{caceres2021probabilistic} on the validity of such approximate approaches from a Bayesian point of view. Thus, MCD is criticized for using a dropout rate that does not depend on the data. Further, \citet{farquhar2020radial} highlights the pathological overconfidence of these methods. This is because they learn an approximate posterior that has discrete support by assigning a zero probability to almost all the neural network weights, which makes them unsuitable as data-dependent prior in continual learning.

\subsection{UQ for deep learning in prognostics}
Most of the recent research in Bayesian deep learning for prognostics is based on the evaluation of only one of the Bayesian deep learning inference techniques. A comparison with other Bayesian deep learning alternatives is very limited or non-existent, and quality of predictive uncertainty is often ignored. In addition, the prominent use of the NASA C-MAPSS \cite{saxena2008damage}, which is a much smaller dataset than N-CMAPSS, hinders scalability assessments. Finally, some of these contributions capture only epistemic uncertainty, either by MCD \cite{peng2019bayesian} or VI \cite{kraus2019forecasting, wang2020recurrent}. Other methods only quantify aleatoric uncertainty by using HNN models whose outputs are the parameters of a likelihood distribution, e.g., $\mu$ and $\sigma$ of a Gaussian distribution \cite{zhao2020probabilistic}. 

We can mention some studies which do quantify both epistemic and aleatoric uncertainties. For instance, VI with Flipout \cite{wen2018flipout} is used with a Gaussian likelihood distribution and RNN architecture \cite{caceres2021probabilistic} to estimate the RUL. The assessment on C-MAPSS, limited to predictive accuracy, show how BNN outperforms MCD and other state-of-the-art deep learning models, especially for the three most complex C-MAPSS subsets. \citet{li2020bayesian} framework uses MCD to capture epistemic uncertainty and several likelihood distributions (Gaussian, logistic, Weibull) to take into account heteroscedastic uncertainty. \citet{benker2021utilizing} compare HMC and Bayes by Backprop \cite{blundell2015weight} performance on C-MAPSS. Aleatoric uncertainty is computed in post-processing in the form of a Gaussian distribution and exploited to further enhanced predictive performance. Also with C-MAPSS, \citet{huang2020bayesian} propose a simple MLP functional model and Bayes by Backprop to quantify epistemic uncertainty, whereas aleatoric uncertainty is estimated by fitting the true RUL values into normal distribution. More recently, a framework \cite{kefalasend2022} tested on C-MAPSS and based on MCD, uses a novel bi-objective Bayesian hyperparameter optimization method, and models the aleatoric uncertainty as a Weibull distribution. A different approach \cite{deng2020controlling} combines a long short-term memory network (LSTM) neural network predictor with a surrogate Wiener propagation model for UQ. 

However, very few comparative studies exist to our knowledge on the performance of UQ in prognostics with Bayesian deep learning. An exception is the recent benchmark by \citet{biggio2021uncertainty}, but the focus is on the evaluation of deep Gaussian processes whose performance is compared only with MCD and MLP. Thus, the study does not cover DE or the recent advances in VI for BNN, and scalability is only tested on a small subset of N-CMAPSS. 

\section{Evaluation framework} \label{sec:framework}
We introduce in this section the UQ methods (\ref{sec:methods}) to be evaluated in our benchmark, including HNN (\ref{sec:heteronn}), several variational BNN techniques (\ref{sec:bnn}), MCD (\ref{sec:mcdropout}) and DE (\ref{sec:deepensembles}). Based on the literature review, these techniques have been selected because of their popularity. In all cases, we assume a Gaussian heteroscedastic setting with a neural network estimating the mean and variance. Also, we describe the experimental setting for the evaluation of these methods, i.e., the N-CMAPSS dataset (\ref{sec:datasets}), models (\ref{sec:models}), evaluation metrics (\ref{sec:metrics}) and uncertainty decomposition techniques (\ref{sec:decomposition}). 

\subsection{Methods} \label{sec:methods}
\subsubsection{Heteroscedastic neural networks (HNN)} \label{sec:heteronn}
Let $X, Y$ be two random variables describing the input and target, following $\prob(X,Y)$. Let $Y$ be conditional on $X$ and assume $\prob(Y \mid X)$ is normally distributed, i.e., $\prob(Y \mid X) = \mathcal{N}(\mu(X),\sigma^2(X))$, where $\mu:\mathbb{R^M}\to \mathbb{R}$ and $\sigma^2:\mathbb{R^M}\to \mathbb{R^+}$ are the true input-dependent mean and variance functions in our Gaussian heteroscedastic setting. A neural network $f_\mathbf{w}$ with weight parameters $\mathbf{w}$ outputs estimates $\hat\mu(X)$ and $\hat\sigma^2(X)$, where the variance is constrained to be positive, which can be implemented for instance with a softplus activation function. 

With a deterministic neural network, $\mathbf{w}$ can be estimated  with MLE by minimizing the NLL loss:

\[
\mathbf{w}^*_{\text{NLL}} = \argminA_\mathbf{w} \mathcal{L}_{\text{NLL}}(\mathbf{w})= \argminA_\mathbf{w} \E_{X,Y} \left [\frac{1}{2}\log\hat{\sigma}^2(X)+\frac{(Y-\hat{\mu}(X))^2}{2\hat{\sigma}^2(X)} + const \right]
\]

\subsubsection{Variational Bayesian neural networks (BNN)} \label{sec:bnn}
In the Bayesian formalism, a prior distribution is defined over the weights of a neural network, and given the data, a posterior distribution is computed to account for the epistemic uncertainty. However, as exact Bayesian inference is computationally intractable for neural networks, variational Bayesian methods have been proposed as an approximate inference method.  

\textbf{Variational inference (VI)}. Let $\prob(\mathbf{w} \mid \mathcal{D})$ be the true posterior distribution over the weights given the training data, $\prob(\mathbf{y} \mid f_\mathbf{w}(x))$ the likelihood function, and $\prob(\mathbf{w})$ the prior. The posterior predictive distribution is computed by marginalization over the weights. At prediction time, the predictive distribution over the target $\mathbf{y^*}$ given a test input $\mathbf{x^*}$ is given by $\prob(\mathbf{y^*} \mid \mathbf{x^*}) = \E_{\prob(\mathbf{w} \mid \mathcal{D})} [ \, \prob(\mathbf{y^*} \mid \mathbf{x^*}, \mathbf{w})] \,$.

This intractable predictive distribution can be approximated by finding a variational posterior distribution \cite{hinton1993keeping, graves2011practical} $q(\mathbf{w} \mid \theta)$ minimizing the Kullback-Leibler (KL) divergence with $\prob(\mathbf{w} \mid \mathcal{D})$:
\[
\theta^* = \argminA_\theta \ \infdiv{q(\mathbf{w} \mid \theta)}{\prob(\mathbf{w} \mid \mathcal{D})}
\]

\paragraph{Mean-field assumption} The mean-field variational inference (MFVI) approach factorizes the variational posterior over its dimensions $q(\mathbf{w} \mid \theta) = \prod_{d=1}^{D} q(\mathbf{w_d} \mid \theta_d)$. This has been used in practice \cite{graves2011practical, blundell2015weight, kingma2015variational} as it is a more computationally tractable setup.

The resulting cost function is known as the evidence lower bound (ELBO) \cite{jaakkola2000bayesian} or  variational free energy \cite{neal1998view}:

\begin{equation}
\label{eqn:elbo}
\mathcal{L}(\mathcal{D}, \theta) = \infdiv{q(\mathbf{w} \mid \theta)}{\prob(\mathbf{w})} - \E_{q(\mathbf{w} \mid \theta)}[ \log \prob(\mathcal{D} \mid \mathbf{w})]
\end{equation}

Note that the first term of (\ref{eqn:elbo}) is the prior-dependent complexity cost, whereas the second one is the data-dependent likelihood cost. The goal is to find a trade-off between the two costs, which is also a way of fitting the data with regularization.

\textbf{Prior distribution choice}. Defining a prior distribution is an important but difficult step in the design of a BNN, because the parameters of an NN do not have a direct interpretation. Nevertheless, the choice of the prior can influence a BNN in several ways: (1) the BNN performance (2) the BNN calibration and (3) how a BNN reacts to OOD samples \cite{silvestro_prior_2020}.

Many priors have been proposed for BNNs \cite{nalisnick_priors_2018}. However, it has been argued that standard Gaussian priors over the parameters are sufficient and the practitioner's beliefs should be represented through the choice of architecture instead \cite{wilson_generalization_2020}. This view had been supported by preliminary studies on small networks and simple problems that did not find conclusive evidence for the misspecification of Gaussian priors \cite{silvestro_prior_2020}. 

Therefore, in our benchmark, we use only Gaussian priors, which have interesting analytical advantages, numerical stability and are already implemented in all BNN libraries.

\textbf{Bayes by Backprop}. As minimizing (\ref{eqn:elbo}) is computationally intractable, \citet{blundell2015weight} proposed a backpropagation-compatible approximation called Bayes by Backprop, which uses Monte Carlo sampling for the expectations. 

\begin{equation}
\mathcal{L}(\mathcal{D}, \theta) \approx \sum_{i=1}^n \log q(\mathbf{w^{(i)}} \mid \theta) - \log \prob(\mathbf{w^{(i)}}) - \log \prob(\mathcal{D} \mid \mathbf{w^{(i)}})
\end{equation}

\noindent where $\mathbf{w^{(i)}}$ denotes the \textit{i}th Monte Carlo sample drawn from the variational posterior, which they assumed to be a diagonal Gaussian distribution (mean-field assumption). 

Let $\mathcal{D}=\{(\mathbf{x_i}, \mathbf{y_i})\}_{i=1}^N$ be the training data with a set of samples drawn from $\prob(X,Y)$. For large datasets $\mathcal{D}$ needing minibatching, a weighting scheme can be applied to the KL term \cite{graves2011practical, blundell2015weight}. If dataset $\mathcal{D}$ is split into $M$ minibatches $\mathcal{D}_1,\ldots,\mathcal{D}_M$ of the same size, let $\pi \in [0,1]^M$ and $\sum_{i=1}^M \pi_i=1$ and define the cost function as:

\begin{equation}
\mathcal{L}_i^\pi(\mathcal{D}, \theta) = \pi_i \infdiv{q(\mathbf{w} \mid \theta)}{\prob(\mathbf{w})}] - \E_{q(\mathbf{w} \mid \theta)}[ \log \prob(\mathcal{D}_i \mid \mathbf{w})
\end{equation}

\noindent with the partitioning scheme $\pi_i = \frac{2^{M-i}}{2^M-1}$, which should lead to faster convergence.

Bayes by Backprop assumes $q(\mathbf{w} \mid \theta)$ can be factorized over the neural network layers $q(\mathbf{w} \mid \theta) = \prod_{l=1}^{L} q_l(\mathbf{w_l} \mid \mathbf{\theta_l})$, where $\mathbf{w_l}$ and $\mathbf{\theta_l}$ denotes respectively the weights and the variational parameters in the layer $l$. We denote a layer forward pass as $\mathcal{F}_{w_l}(x_l)$, with $x_l$ referring to the layer input data. A neural network can then be represented as a composition of layers and activations: $f_w = \mathcal{F}_{w_L} \circ a_{L-1} \circ \mathcal{F}_{w_{L-1}}, \ldots, \circ ~a_1 \circ \mathcal{F}_{w_1}$, where $a_{l}$ denotes the layer activation function.

\textbf{Gradient estimation with naive reparametrization}. Gradients $\nabla_\theta \mathcal{L}(\mathcal{D}, \theta)$ can be estimated by sampling weights $\mathbf{w_l}$ to compute $\mathcal{F}_{w_l} (x_l)$ for each layer with $\mathbf{w}_l = g(\theta_l, \epsilon_l)$, e.g. $\mathbf{w}_l=\mu_l + \sigma_l \odot \epsilon_l$. The problem with this naive reparametrization is the high variance of the resulting gradients, which negatively impacts the performance of the gradient descent. This is because $\mathbf{w_l}$ is shared with all the inputs $x_i$ in a minibatch $D_i$ introducing correlation between the terms $\nabla_\theta \log \prob(\mathbf{y}_i \mid f_\mathbf{w}(\mathbf{x}_i))$. This correlation is zero when a separate sample is drawn for each input $x_i$ in the batch, but this leads to a prohibitive memory cost of $\mathcal{O}(N_{l, in}N_{l,out}|\mathcal{D}_i|)$, i.e., proportional to the number of parameters in the layer, $N_{l, in}, N_{l,out}$. 

\textbf{The local reparametrization trick (LRT)}. This technique is proposed by \citet{kingma2015variational} to reduce gradient variance. It can only be applied to linear neural network layers under the mean-field assumption, i.e., it is not valid for layers with weight sharing, such as convolutional or recurrent layers. LRT is based on sampling $\mathcal{F}_{w_l} (x_l)$ instead of the weights $\mathbf{w}_l \sim q(\cdot \mid \theta)$, with the following reparametrization for the forward pass: 

\begin{equation}
\label{eqn:lrt}
\mathcal{F}_{w_l} (x_l) = \mathcal{F}_{\mu_l} (x_l) + \epsilon_l \odot \sqrt{\mathcal{F}_{\sigma^2_l} (x_l^2)}
\end{equation}

\noindent where $\mathbf{w}_l \sim  \mathcal{N}(\mu_l,diag[\sigma_l^2])$, $\epsilon_l \sim \mathcal{N}(\mathbf{0},\mathbf{I}_l)$ and $\odot$ denotes elementwise multiplication. According to Equation \ref{eqn:lrt}, LRT requires two forward passes, but the memory cost is reduced by a factor of $N_{l,in}$ to $\mathcal{O}(N_{l,out}|\mathcal{D}_i|)$.

\textbf{Flipout (FO)}. This technique by \citet{wen2018flipout} is an alternative to LRT for gradient variance reduction under the mean-field assumption, which works with convolutional and recurrent layers in addition to fully-connected layers. It decorrelates variational samples among inputs in the minibatch by exploiting the algebraic property $(\mathbf{w}_l \odot \epsilon_1 \epsilon_2^T) x_l = \epsilon_1 \odot \mathbf{w}_l (x_l \odot \epsilon_2)$, where $\epsilon_1, \epsilon_2$ are random vectors sampled uniformly from $\pm 1$.

\textbf{Radial BNN (RAD)}. \citet{farquhar2020radial} propose a solution to make MFVI scale to larger models by addressing the sampling problem with Gaussian variational posteriors in high dimensions. This phenomenon, known as the `soap-bubble' ~\cite{bishop2006pattern}, refers to the probability mass of high-dimensional multivariate Gaussian being clustered in a narrow shell at a radius depending on both the variance and the number of dimensions. Samples from such posterior distributions tend to be very far away from the mean, which leads to high gradient variance, making optimization of large MFVI models difficult. The proposed solution consists of a small modification of the weight sampling by dividing the MFVI noise samples by their norm and multiplying it by a distance sampled as a centered Gaussian:

Let $r = |\hat{r}|,\ \hat{r} \sim \mathcal{N}(0,1) $ and $ \epsilon \sim \mathcal{N}(0, 1)$. The two methods sample their weights according to:
\begin{align*} 
\text{MFVI} &:  \omega := \mu + \sigma \odot \epsilon \\
\text{Radial} &: \omega := \mu + \sigma \odot \frac{\epsilon}{||\epsilon||} * r
\end{align*} 

\subsubsection{MC Dropout (MCD)}\label{sec:mcdropout}
Dropout is originally a regularization technique used during the training phase of neural networks \cite{srivastava2014dropout}. More recently, \citet{gal_dropout_2016} proposed MCD as a method that uses dropout at test time to estimate predictive uncertainty. In the literature, MCD has been interpreted in terms of both Bayesian approximation \cite{gal_dropout_2016, kingma2015variational} and ensemble model combination \cite{srivastava2014dropout}. MCD is a VI technique, with variational distribution  $q(\mathbf{w}_l \mid \theta_l)$ defined as:

\begin{equation}
\label{eq:mcdropout}
\begin{aligned} 
\mathbf{w}_l &= \theta_l \odot diag(\epsilon_l) \\
\epsilon_l &\sim Bernouilli(p_l)
\end{aligned} 
\end{equation}

\noindent where $\theta_l$ denotes the weight parameter matrix for $l$th level of dimensions $N_{l, in} \times N_{l, out}$, and the binary vector $\epsilon_l$ corresponds to the units in layer $l-1$ to be dropped out as an input to layer $l$.

Let $E(\cdot,\cdot)$ be a loss function, e.g., MSE. When a neural network is trained with a $L_2$ regularization term, the objective function:

$$
\mathcal{L}_{dropout} = \frac{1}{N} \sum_{i=1}^N E(y_i, y_i^*) + \lambda \sum_{l=1}^L( || \mathbf{w}_l ||_2^2) 
$$

\noindent is equivalent to the ELBO for VI, assuming a normal prior on the weights and the distribution presented in Equation \ref{eq:mcdropout} as variational posterior.
 
\subsubsection{Deep ensembles (DE)} \label{sec:deepensembles}
\citet{lakshminarayanan_simple_2017} developed this technique as an alternative to MCD for predictive uncertainty estimation. By initializing an ensemble of networks with different random seeds and training them on the same dataset, their predictions are expected to be similar, while not being exactly the same. However, DE comes at the cost of training multiple models and multiplying the number of parameters, so it is less computationally efficient than MCD. 

In the context of a supervised regression problem, a mean $\mu$ and a variance $\sigma^2$ are computed independently by each neural network in the ensemble. For an input $x_i$ of the dataset, the mixture mean and variance are computed as follows:
$$ \hat{\mu}(x_i) = \frac{1}{M}\sum_{m=1}^M \mu_m(x_i) $$
$$ \hat{\sigma}^2(\mathbf{x_i}) = \frac{1}{M}\sum_{m=1}^M \left(\sigma_m^2(\mathbf{x_i}) + \mu_m^2(\mathbf{x_i})\right)- \mu^2(\mathbf{x_i}) $$ 

\subsection{Datasets} \label{sec:datasets}
We tested the different UQ methods on the new C-MAPSS (N-CMAPSS) dataset for aircraft turbofan engines created by \citet{arias2021aircraft}. This dataset improves the degree of fidelity of the popular NASA C-MAPSS dataset \cite{saxena2008damage} and contains run-to-failure trajectories of sensor data generated by the Commercial Modular Aero-Propulsion System Simulation (C-MAPSS) simulation software. 

In C-MAPSS, simulated flight conditions (only standard cruise) and failure modes are limited. Also, the onset of abnormal degradation is independent of the past operating profile, resulting in a lack of realism in the simulated degradation trajectories. In N-CMAPSS, however, the full history of engine degradation is simulated from recorded onboard data of a commercial jet, covering all flight phases and more routes than C-MAPSS. Sensor data is sampled at $1$\,Hz, which makes N-CMAPSS a large dataset (14\,GB) adapted for the benchmark of deep learning models.

\subsubsection{Dataset selection}
From the nine subsets available in N-CMAPSS, we selected the first five (\texttt{D1} to \texttt{D5}) for our benchmark, which contains the run-to-failure trajectories of 54 turbofan units. There are 33 units in the development set (used for training and validation) and 21 in the testing set. Please refer to \cite{arias2021aircraft} for a detailed explanation of failure mode simulation for the five components of the turbofan, i.e., fan, low-pressure compressor (LPC), high-pressure compressor (HPC), low-pressure turbine (LPT) and high-pressure turbine (HPT), which can be affected by degradation in flow (F) and efficiency (E). Flight class (FC) corresponds with short (1), medium (2), and long haul flight (3) categories. The distribution of flight classes and failure modes per dataset is provided in Table \ref{tab:datasets_detail}.

\begin{table}[t!]
    \centering
    \begin{tabular}{cccccccccc}
        \hline
        \textbf{\thead{DS}} &  \textbf{\thead{FC}} &  \textbf{\thead{Fan}} &  \textbf{\thead{HPC}} & \textbf{\thead{HPT}} & \textbf{\thead{LPC}} &  \textbf{\thead{LPT}} \\
        \hline
         1 & 1,2,3 &  - &   - &   E &   - &   - \\
         2 & 1,2,3 &  - &   - &   E &   - & E,F \\
         3 & 1,2,3 &   - &   - &   E &   - & E,F \\
         4 & 2,3   &  E,F &   - &   - &   - &   - \\
         5 & 1,2,3 &  - & E,F &   - &   - &   - \\
        \hline
    \end{tabular}
    \caption{Distribution of flight classes and failure modes simulated in the datasets.}
    \label{tab:datasets_detail}
\end{table}

One of the goals of our benchmark is to evaluate the performance of the methods with complex or OoD data. The N-CMAPSS dataset enables such evaluation since the distributions of measured features $X_s$ for training and test set on each of the data subsets can be notably different for two reasons: 

 \textit{Different flight conditions, $W$}. The assignation of one of the three possible flight classes to each unit in the training and test sets is random. Therefore, the training and test sets contain data from the flight classes in different proportions (see Table \ref{tab:datasets}). Since each flight's classes involve notably different flight conditions (e.g., flight altitudes), the training set is imbalanced with respect to the flight conditions on the test set.

 \textit{Different degradation trajectories, $\theta(t)$}. As documented in \cite{arias2021aircraft}, the degradation effect on the engine components was simulated in CMAPSS by modifying health-related parameters $\theta$ (i.e., the causal factors of degradation), and imposing a different (stochastic) evolution of these parameters over time for each unit, i.e., a degradation trajectory. As a result, some units experience a fast degradation and others a slow one. Moreover, in most data subsets, the degradation involves the simultaneous modification of several model health parameters affecting the flow capacity and efficiency of the five main components of the turbofan. Therefore, test units can exhibit degradation trajectories involving extrapolation of the health parameters, hence creating a potential "out of distribution (OOD)" scenario. 

\begin{figure}[!]
    \centering
    \includegraphics[scale=0.4]{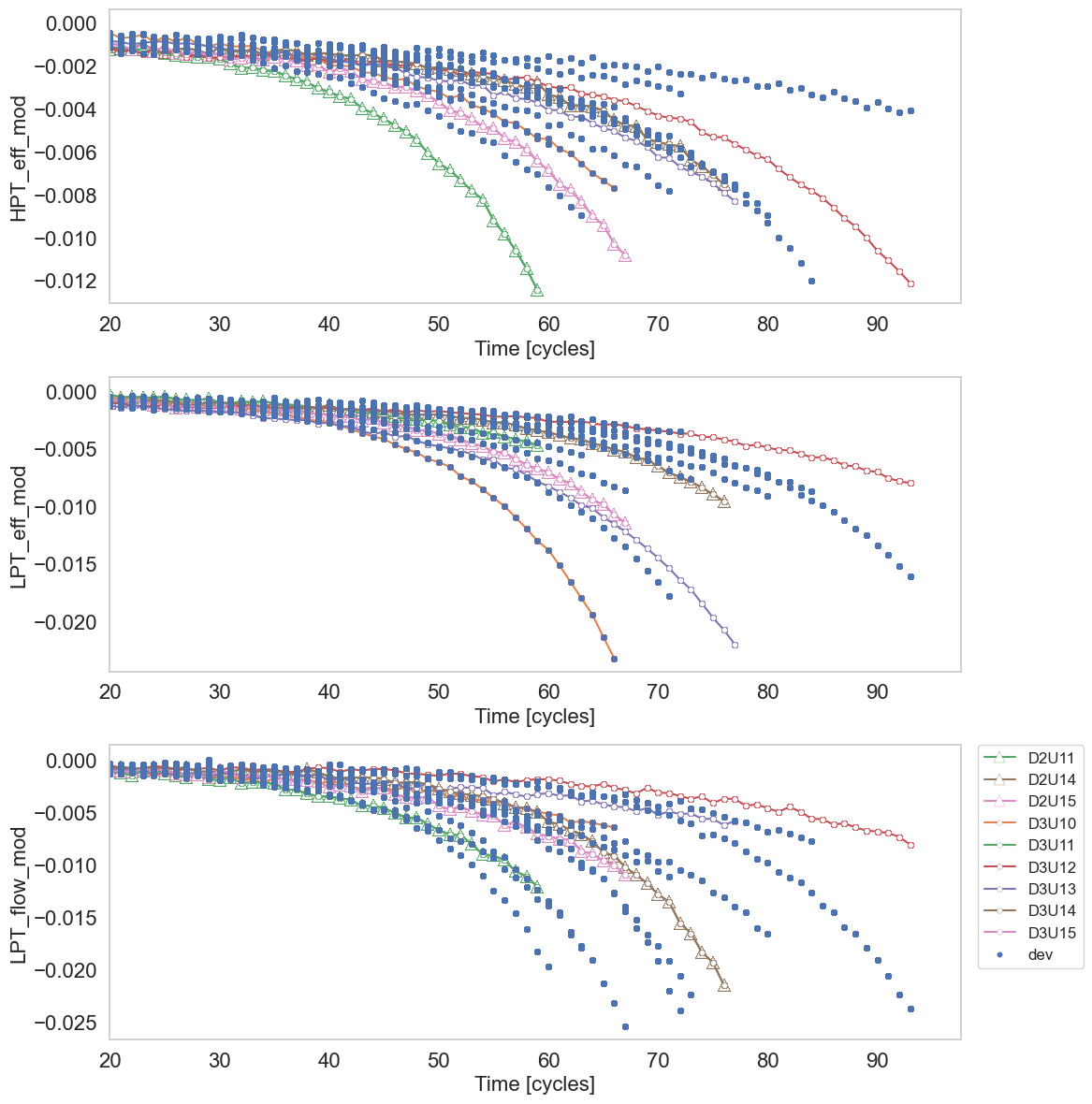}
    \caption{Evolution of the three causal factors of degradation $\theta$ = \texttt{\{HPT\_eff\_mod, LPT\_flow\_mod, LPT\_Eff\_mod\}} with time for development (blue), and the nine test units in \texttt{D2} and \texttt{D3}. \texttt{D2U11} (green triangles), \texttt{D3U11} (green dots) and \texttt{D3U12} (red dots) are OOD cases since they involve extrapolations of $\theta$ outside the development data.}
    \label{fig:theta_vs_cycle}
\end{figure}

\begin{figure}[!]
    \centering
    \includegraphics[scale=0.55]{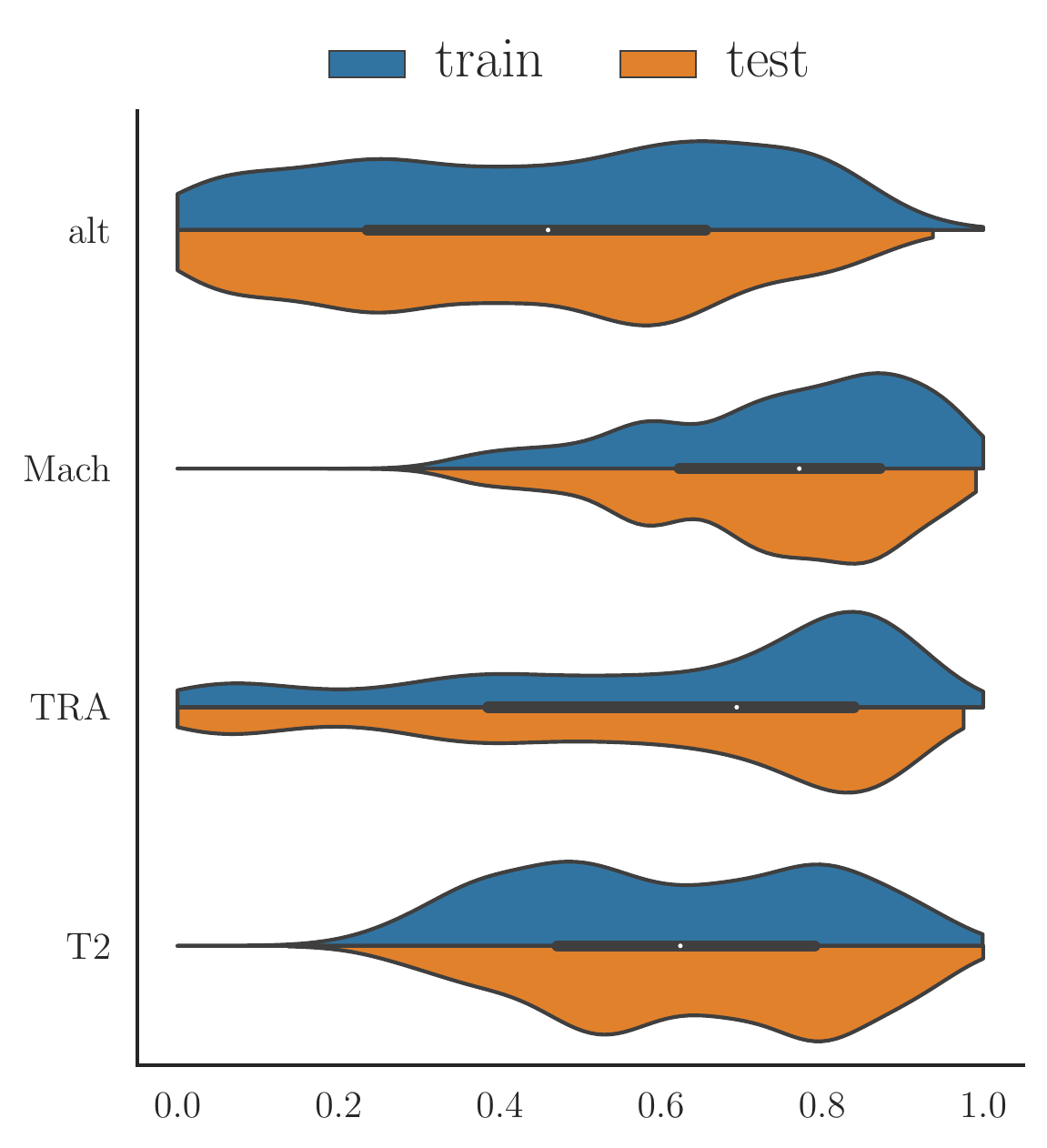}
    \includegraphics[scale=0.35]{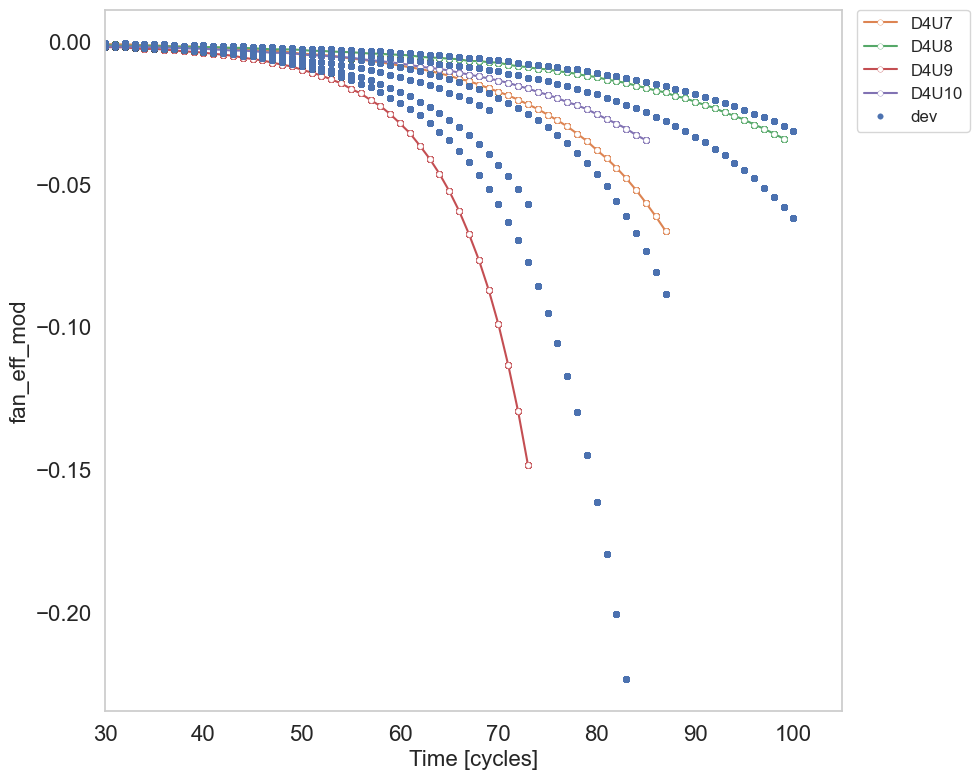}
    \caption{(Left) Comparison between the train and test distributions of W parameters for units in \texttt{D4}. (Right) The degradation trajectory of Unit \texttt{D4U9} (red) involves extrapolation on \{fan\_eff\_mod\}.}
    \label{fig:ds4_wdist_rul}
\end{figure}

Such an OOD scenario can be observed for several units within data subsets \texttt{D2}, \texttt{D3}, and \texttt{D4}. In particular, units in data subsets \texttt{D2}, \texttt{D3} are affected by a failure mode that involves the degradation of the low-pressure turbine (LPT) efficiency and flows in combination with the high-pressure turbine (HPT) efficiency, i.e., engine health parameters $\theta$. Figure \ref{fig:theta_vs_cycle} shows the degradation trajectories for development (blue), and the nine test units in \texttt{D2} and  \texttt{D3} i.e., the evolution of the three causal factors of degradation $\theta$ = \texttt{\{HPT\_eff\_mod, LPT\_flow\_mod, LPT\_Eff\_mod\}} with time. We can observe that \texttt{D2U11} (green triangles), \texttt{D3U11} (green dots), and \texttt{D3U12} (red dots) follow degradation trajectories that involve extrapolations outside the development data. As depicted in Figure \ref{fig:ds4_wdist_rul} (right), a similar OOD behavior can be observed \texttt{D4U09}. 

It is worth mentioning that among the five considered datasets, \texttt{D4} is a specially challenging dataset because of a couple of specificities (see Table \ref{tab:datasets_detail}): 1) it is the only dataset covering the degradation of the fan component, 2) it only contains flight classes 2 and 3. However, the distribution of the operating \texttt{W} parameters in the training and test sets are close enough (see the left plot in Figure \ref{fig:ds4_wdist_rul}). 

\subsubsection{Feature selection and pre-processing}
We selected the N-CMAPSS scenario descriptors and sensor measurements described in Table \ref{tab:features} as inputs, as well as the target RUL for our supervised learning approach. This makes an input size of $4 + 14 = 18$ values per time step. 

\begin{table}[h!]
    \centering
    \begin{tabular}{ll}
        \hline
        Features & Description  \\
        \hline
        $W$ & 4 scenario descriptors (flight conditions): altitude, Mach, TRA, T2 
        \\
        $X_s$ & 14 sensor measurements: physical fan speed, temp. at LPC outlet, ... \\
        Y & RUL label (remaining flight cycles) \\
        \hline
    \end{tabular}
    \caption{N-CMAPSS selected features.}
    \label{tab:features}
\end{table}

We used a sliding window of length $30$ over the time series, so the resulting input size is $18 \times 30$. To speed up computation, we downsampled sensor data to $0.1$\,Hz by decimation, which results in windows of 5 minutes. All the input features were standardized to have zero mean and one standard deviation. We randomly split the sliding windows in the development set into 90\% for training and 10\% for validation. The number of units and window samples per dataset is shown in Table \ref{tab:datasets}. 

\begin{table}[h!]
    \centering
    \begin{tabular}{lrrrrr}
        \hline
        Dataset & \# Units & \# Samples & FC1 & FC2 & FC3 \\
        \hline
        Train & 33 & $238\,182$ & $0.10$ & $0.27$ & $0.62$ \\
        Validation & 33 & $26\,464$ & $0.11$ & $0.27$ & $0.61$ \\
        Test & 21 & $144\,015$ & $0.17$ & $0.34$ & $0.49$\\
        \hline
    \end{tabular}
    \caption{Number of engine units, sliding window samples and flight class ratios per dataset.}
    \label{tab:datasets}
\end{table}

\subsubsection{Piece-wise linear RUL degradation}
The ideal true RUL is linearly decreasing over time, but in N-CMAPSS, engines follows a two-stage degradation trend. They stay healthy during the first cycles of operation. Then, their initial healthy state gradually deteriorates over the operation until reaching the end of life. Therefore, we could have used the piece-wise linear RUL target label function proposed in \cite{zhang2018long} for C-MAPSS to limit the maximum value of RUL to a constant value for all engines. However, setting a maximum RUL label for each engine is suboptimal, since each engine has a different life cycle. Instead, we corrected the labeling with the maximum RUL of each engine set to the RUL at the moment when a fault occurs and its actual deterioration begins, which is given by the parameter $h_s$ included in the N-CMAPSS auxiliary data.

\subsection{Model architectures} \label{sec:models}
\subsubsection{Functional models}
Functional models are deep neural network architectures used to build the probabilistic neural networks. After some preliminary testing with an adaptation of a fully-connected model (\texttt{D3} \cite{benker2021utilizing}) and a convolutional one (\texttt{C2P2} \cite{sateesh2016deep}) originally designed for C-MAPSS, we selected the \texttt{Inception} model \cite{devol2021inception} as it clearly outperformed the others. 

\begin{figure}
    \centering
    \includegraphics[scale=0.65]{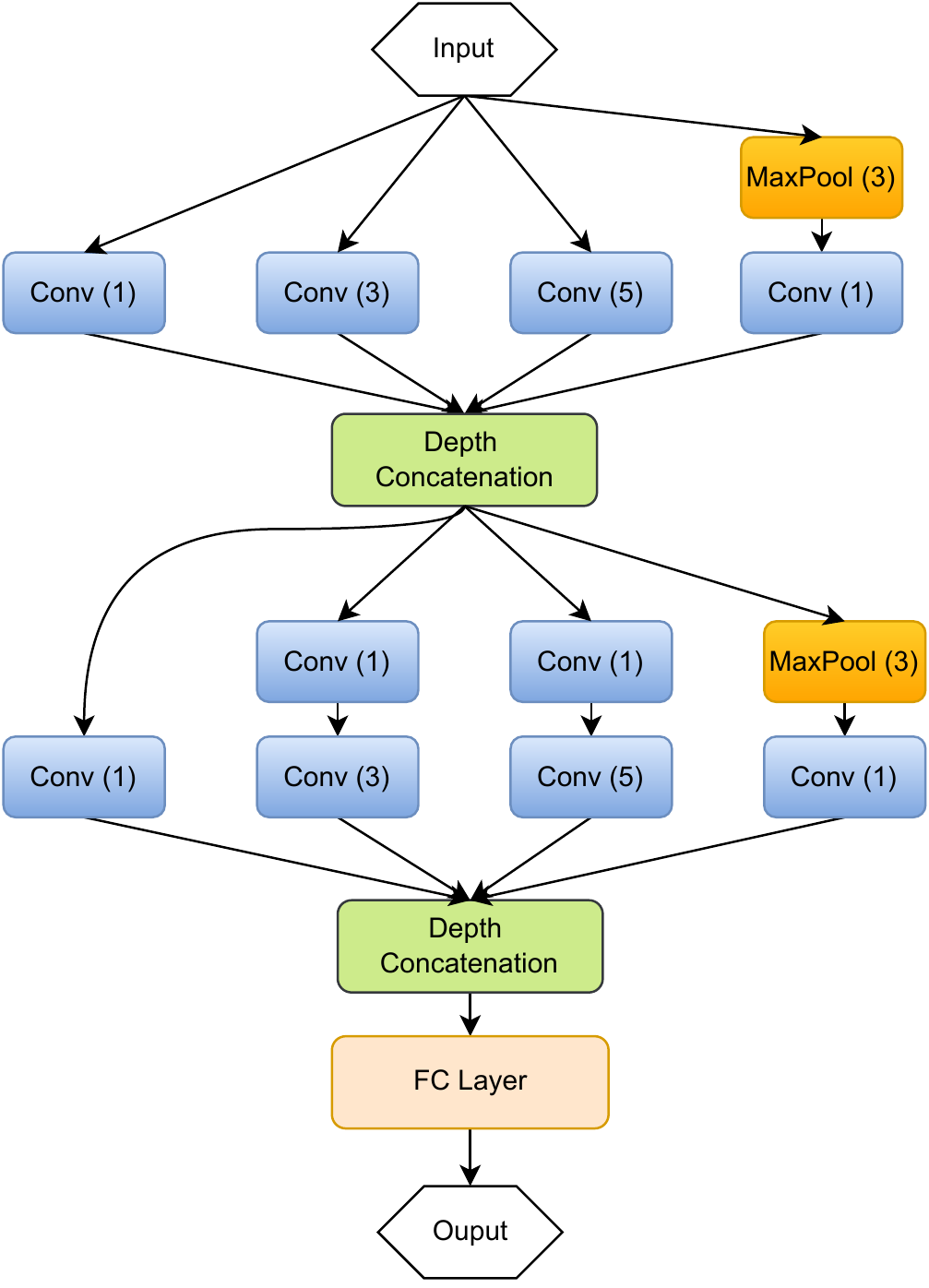}
    \caption{Inception model by DeVol et al. \cite{devol2021inception}}
    \label{fig:inception_model}
\end{figure}

The \texttt{Inception} architecture (see Figure \ref{fig:inception_model}) achieved the second-best performance in the 2021 PHM Data Challenge to find the best deep learning architecture for N-CMAPSS. It uses several inception modules \cite{szegedy2015going} to capture information lying at different scales through convolutions of different sizes applied to the same input. We used \texttt{ReLU} as activation function. The number of parameters of this model is $\sim 187k$, which will be doubled when turned into a BNN. 

The functional models were implemented with \texttt{PyTorch} and \texttt{PyTorch Lightning} \cite{Falcon_PyTorch_Lightning_2019}, and turned automatically into BNN with \texttt{TyXe} \cite{ritter2022tyxe} based on \texttt{Pyro} \cite{bingham2019pyro}. 

\subsubsection{Hyperparameter search}
Hyperparameter tuning is particularly important for inference methods like MFVI suffering from hyperparameter sensitivity \cite{farquhar2020radial}. We implemented the hyperparameter search with \texttt{Optuna} \cite{optuna_2019} and performed 60 trials for each method. Models were trained with the Adam optimizer for only 20 epochs per trial to speed up the process. The hyperparameter search took about 5 days on 3 \textit{NVIDIA GeForce GTX 1080 Ti} GPUs. We chose a sampling algorithm called TPESampler \cite{bergstra2011algorithms} combined with the median pruner algorithm available in \texttt{Optuna}.

For the BNN techniques, the search space was guided by the validation ELBO (20 Monte-Carlo (MC) samples of $q_\theta$ were used for validation) and included the exploration of the following parameters:

\begin{itemize}
    \item Functional model pre-training with root-mean-square error (RMSE) (0 or 5 epochs)
    \item Number of $q_\theta$ MC samples for estimating training loss (1 or 2 samples)
    \item Gaussian prior scale (log-uniformly sampled between $10^{-2}$ and $0.5$)
    \item Gaussian $q_\theta$ scale (log-uniformly sampled between $10^{-4}$ and $10^{-2}$)
    \item The learning rate (log-uniformly sampled between $10^{-5}$ and $10^{-3}$)
    \item Batch size (100, 250, 500 or 1000 samples)
\end{itemize}

The pretraining of the functional model for a few epochs is a technique to help initialize the means of the Normal weight distributions. The best BNN hyperparameters found are listed in Table \ref{tab:bnn_params}.

\begin{table}[b]
    \centering
    \begin{tabular}{lrrr} \hline
         & LRT & FO & RAD \\ \hline
        Training MC samples & 1 & 2 & 1 \\
        Pre-train epochs & 5 & 5 & 5 \\ 
        Prior scale & 0.138793 & 0.198768 & 0.092516 \\ 
        $q_\theta$ scale & 0.001351 & 0.000214 & 0.001241 \\
        Learning rate & 0.000857 & 0.000948 & 0.000956 \\
        Batch size & 100 & 100 & 100 \\
        \hline
    \end{tabular}
    \caption{Hyperparameters for the variational BNN techniques.}
    \label{tab:bnn_params}
\end{table}

The search space for the other techniques was guided by the validation NLL loss. For HNN, it included the batch size (same choices as with BNN) and the learning rate (log-uniformly sampled between $10^{-4}$ and $5^{-3}$). For MCD, we had, in addition, the number of MC samples (20, 50, 100) and dropout probability (uniformly sampled between $0.20$ and $0.85$). The final hyperparameters for these methods are listed in Table \ref{tab:nn_params}.   

\begin{table}
    \centering
    \begin{tabular}{lrr} \hline
         & HNN & MCD \\ \hline
        MC samples & - & 100 \\
        Dropout probability & - & 0.241437 \\
        Learning rate & 0.001574 & 0.000772 \\
        Batch size & 250 & 100 \\
        \hline
    \end{tabular}
    \caption{Hyperparameters for HNN and MCD.}
    \label{tab:nn_params}
\end{table}

\subsubsection{Model training procedure}

We performed 10 runs for HNN and 5 runs for the other methods by using different random seeds. The number of HNN runs was twice as many because we needed enough independently trained HNN models to build the DE. Indeed, there was no proper training involved for DE models. Instead, we generated 5 DE models as random combinations of 5 out of the 10 available HNN models. More than 5 models in the ensemble could have possibly resulted in better performance, but also in an unfair comparison with the other methods because of the huge difference in the number of parameters. 
We trained the models for a maximum number of 500 epochs using early stopping with patience 50 (HNN and MCD) and 20 (BNN). All models were trained with the best hyperparameters listed in Table \ref{tab:bnn_params} and Table \ref{tab:nn_params}. We minimized the validation ELBO for the BNNs, and the validation NLL for MCD and HNN. 

\begin{figure}
    \centering
    \includegraphics[scale=0.5]{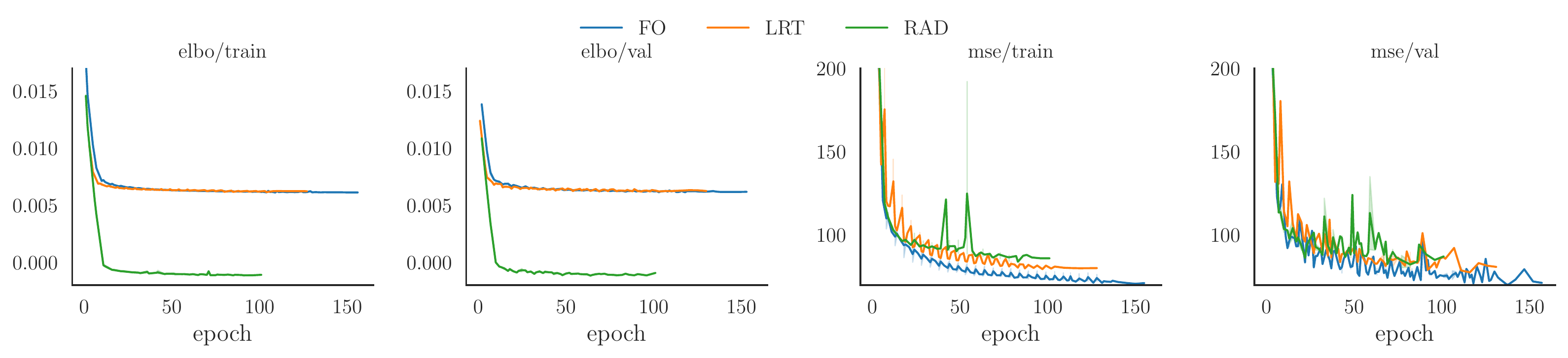}
    \caption{Losses during training and validation of the BNN methods.}
    \label{fig:bayesian_loss}
\end{figure}

Figure \ref{fig:bayesian_loss} plots train and validation ELBO and MSE losses for the BNN models. Training and validation curves are very close, especially in the case of the ELBO. Early stopping is activated earlier for RAD, whose validation ELBO converges faster and at a lower level than the LRT and FO ones. In terms of the validation MSE, however, the RAD curve is noisier and converges to a higher MSE value.   

\begin{figure}
    \centering
    \includegraphics[scale=0.5]{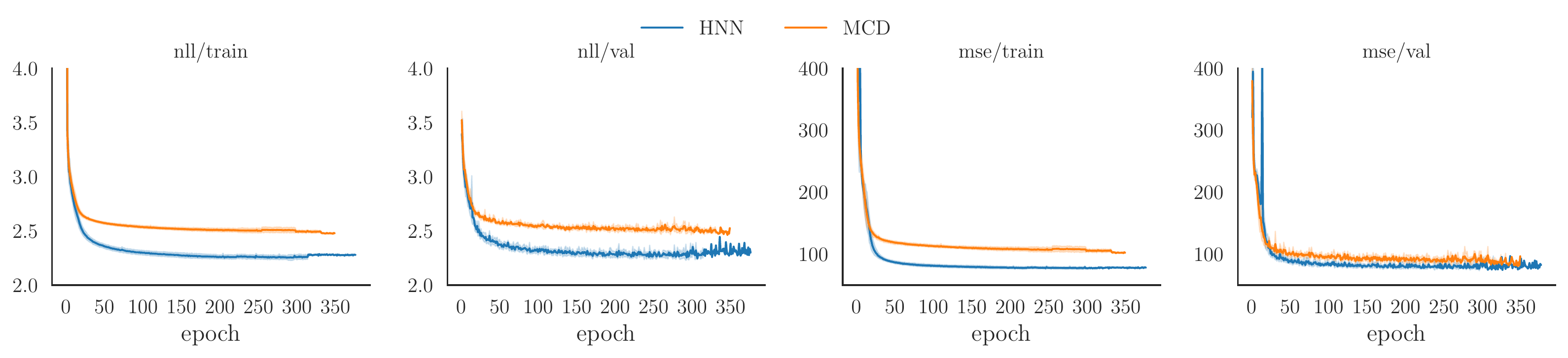}
    \caption{Losses during training and validation of HNN and MCD methods.}
    \label{fig:freq_loss}
\end{figure}

Training and validation losses for the HNN and MCD methods are shown in Figure \ref{fig:freq_loss}. We discarded 2 out of the 10 HNN runs because of significant anomalies in the training and validation curves. Otherwise, all runs produced very similar training and validation curves, with MCD converging slightly before HNN but at a higher loss level.

Table \ref{tab:run_stats} provides some statistics showing big differences among the methods in terms of epochs and hours needed for each run. BNN models converge much faster than HNN and MCD, but the time taken to complete a run is significantly higher. Among the BNN techniques, the run time of RAD is the lowest and similar to the MCD one, whereas the one of FO is by far the highest of all methods. HNN is the fastest method and the one with the highest variance in the number of epochs. The latter is due to the two discarded runs which were stopped much earlier by the early stopping mechanism.

\begin{table}
    \centering
    \begin{tabular}{lcc}
        \hline
         & Epochs & Time (h) \\
        \hline
        RAD & 63.0 $\pm$ 15.2 & 6.8 $\pm$ 1.2 \\
        LRT & 80.6 $\pm$ 19.6 & 8.7 $\pm$ 1.6 \\
        FO & 116.4 $\pm$ 13.4 & 20.6 $\pm$ 1.8 \\
        HNN & 177.5 $\pm$ 107.2 & 1.4 $\pm$ 0.7 \\
        MCD & 241.0 $\pm$ 52.5 & 6.2 $\pm$ 0.9 \\
        \hline
    \end{tabular}
    \caption{Number of epochs needed for a run to converge to the lowest validation loss and the number of hours taken for a run to finish with early stopping. Rows are sorted by the average number of epochs.}
    \label{tab:run_stats}
\end{table}

\subsection{Evaluation metrics} \label{sec:metrics}
We evaluated model accuracy with mean absolute error (MAE), root-mean-square error (RMSE) and NLL. MAE is less sensitive to outliers than RMSE. In fact, RMSE provides a measure of the worst-case accuracy. NLL is a proper scoring rule \cite{gneiting2007strictly} assessing both the predictive accuracy of a model and the quality of its UQ estimate \cite{tran2020methods}. However, NLL can over-emphasize tail probabilities \cite{ovadia2019can} and favors models that tend to be under-confident rather than over-confident \cite{quinonero2005evaluating}). We further assessed predictive uncertainty by measuring calibration and sharpness \cite{kuleshov2018accurate} with the tools provided in the \texttt{Uncertainty Toolbox} \cite{chung2021uncertainty}.

Calibration can be visualized with a calibration curve, which \lq displays the true frequency of points in each interval relative to the predicted fraction of points in that interval\rq~\cite{kuleshov2018accurate}. Thus, for $m$ confidence levels $0 \leq p_1 < p_2 < \ldots < p_m \leq 1$ and a test set $\mathcal{T}=\{x_t, y_t\}_{t=1}^T$, we can compute the empirical frequency for each threshold $p_j$ as:

\[
\hat{p}_j = \frac{|\{ y_t \mid F_t(y_t) \leq p_j, t=1, \ldots,T \}|}{T}
\]

\noindent where $F_t(y_t \mid \hat{\mu}(x_t), \hat{\sigma}^2(x_t) )=\frac{1}{2}\left[1+\mathrm{erf}\left(\frac{y_t-\hat{\mu}(x_t)}{\hat{\sigma}^2(x_t){\sqrt{2}}}\right)\right]$ is the Gaussian CDF. The calibration curve is built from $\{(p_j,\hat{p}_j)\}_{j=1}^m$ and the root mean square calibration error (RMSCE) is computed as:

\begin{equation}
\mathrm{RMSCE} (F_1,y_1,\ldots,F_T,y_T) = \sqrt{\frac{1}{m} \sum_{j=1}^m (p_j-\hat{p}_j)^2} 
\end{equation}

Well-calibrated models can still have large uncertainty estimates, which are less informative than smaller ones. We assess this aspect with the sharpness score \cite{kuleshov2018accurate}, which is simply the average of the predicted variances:
\begin{equation}
\mathrm{Sharp}(F_1,\ldots,F_T) = \sqrt{\frac{1}{T} \sum_{t=1}^T \hat{\sigma}^2(x_t)}
\end{equation}

Please note that unlike in  \cite{kuleshov2018accurate}, we added the square root as suggested in \cite{tran2020methods} to give sharpness the same units as the predictions.

\subsection{Uncertainty decomposition} \label{sec:decomposition}
We assessed for the different methods the possibility of separating the epistemic and aleatoric uncertainty from the total predictive uncertainty. In VI, we can derive this decomposition from the Bayesian Model Average of the variational approximation by adapting the method in \cite{kwon_uncertainty_2020}. The posterior predictive distribution is defined as: 
$$ 
p(y^*|x^*, D) = \int_\Omega p(y^*|x^*, \omega)p(\omega|D)d\omega \approx \int_\Omega p(y^*|x^*, \omega) q_\theta(\omega) d\omega  
$$

\noindent We can then use it in the next equations
$$
\begin{aligned}
\mathop{\mathbb{E}\left[ y^* \right]}_{p(y^*|x^*, D)} & = \int y^* p(y^*|x^*, D) dy^* \\
 & = \int_\Omega q_\theta(\omega) \int y^* p(y^*|x^*, \omega) \ d y^* d\omega \\
 & = \int_\Omega q_\theta(\omega) \mathop{\mathbb{E}\left[ y^* \right]}_{p(y^*|x^*, \omega)} d\omega \\
 & = \mathop{\mathbb{E}}_{ \omega \sim q_{\theta}} \left[  \mathop{\mathbb{E}\left[ y^* \right]}_{p(y^*|x^*, \omega)}  \right]
\end{aligned}
$$
In the same manner,
$$
\begin{aligned}
\mathop{\mathbb{E}\left[ (y^*)^2 \right]}_{p(y^*|x^*, D)} & = \int_\Omega q_\theta(\omega) \int y^{*2} p(y^*|x^*, \omega) \ d y^* d\omega \\
& = \int_\Omega q_\theta(\omega) \mathop{\mathbb{E}\left[ y^{*2} \right]}_{p(y^*|x^*, \omega)} d\omega \\
& = \int_\Omega q_\theta(\omega) \left( 
    \mathop{\text{Var}}_{p(y^{*}|x^*, \omega)}(y^*) + \mathop{\mathbb{E}\left[y^{*}\right]^2}_{p(y^{*}|x^*, \omega)}
\right) d\omega  \\
\end{aligned}
$$
To compute the total variance
$$
\begin{aligned}
\mathop{\text{Var}}_{p(y^{*}|x^*, D)}(y^*) & = \mathop{\mathbb{E}\left[ (y^{*})^{2} \right]}_{p(y^*|x^*, D)} - \left( \mathop{\mathbb{E}\left[ (y^*) \right]}_{p(y^*|x^*, D)} \right)^2 \\
& = \int_\Omega q_{\theta}(\omega) \left( \mathop{\text{Var}}_{p(y^{*}|x^*, \omega)}(y^*) + \mathop{\mathbb{E}\left[y^{*}\right]^2}_{p(y^{*}|x^*, \omega)} \right) d\omega \  - \left( \int_\Omega q_{\theta}(\omega) \mathop{\mathbb{E}\left[ y^{*} \right]}_{p(y^*|x^*, \omega)} d\omega \right)^2  \\
& = \mathop{\mathbb{E}}_{\omega \sim q_{\theta}(\omega)} \left[ \mathop{\text{Var}}_{p(y^*|x^*, \omega)}(y^*) \right] + \int_\Omega q_{\theta}(\omega) \mathop{\mathbb{E}\left[ y^* \right]^2}_{p(y^*|x^*, \omega)} d\omega  - \left( \int_\Omega q_{\theta}(\omega) \mathop{\mathbb{E}\left[ y^{*} \right]}_{p(y^*|x^*, \omega)} d\omega \right)^2 \\
& = \mathop{\mathbb{E}}_{\omega \sim q_{\theta}(\omega)} \left[ \mathop{\text{Var}}_{p(y^*|x^*, \omega)}(y^*) \right] + \mathop{\text{Var}}_{\omega \sim q_{\theta}(\omega)}\left[ \mathop{\mathbb{E}\left[ y^* \right]}_{p(y^*|x^*, \omega)} \right]
\end{aligned}
$$
Which is the analytical formula for
\begin{equation} \label{equ:ep_al_unc_analytical}
    \mathop{\text{Var}}_{p(y^{*}|x^*, D)}(y^*) = \mathop{\text{Var}}_{\omega \sim q_{\theta}(\omega)}\left( \hat{\mu} \right) + \mathop{\mathbb{E} \left[ \hat{\sigma}^2 \right]}_{\omega \sim q_{\theta}(\omega)}
\end{equation}  

For a BNN or with MCD, the total predictive variance for outputs $\hat{\mu}$ and $\hat{\sigma}$ is computed by Monte-Carlo. Let vectors $\hat{\mu} = (\hat{\mu}_1, ..., \hat{\mu}_n)$ and $\hat{\sigma} = (\hat{\sigma}_1, ..., \hat{\sigma}_n)$ be the results of $n$ forward passes in the BNN with different weight samplings. Then, the total uncertainty $\hat{\sigma}_{\text{tot}}^2$ is defined as:
\begin{equation} \label{equ:ep_al_unc}
    \hat{\sigma}_{\text{tot}}^2(x) = \underbrace{\frac{1}{N}\sum_{i=1}^N \hat{\mu}_i(x)^2  - \left(\frac{1}{N} \sum_{i=1}^{N} \hat{\mu}_i(x)\right)^2}_{\text{epistemic part}} + \underbrace{\frac{1}{N} \sum_{i=1}^{N} \hat{\sigma}_i(x)^2}_{\text{aleatoric part}}
\end{equation}  

Concerning DE, the decomposition of predictive uncertainty is less straightforward. Recently, Egele et al. \cite{egele2021autodeuq} proposed a method based on the law of total variance \cite{egele2021autodeuq}, where they randomly select $K$ models from a catalog to form the ensemble. However, they assume a great model diversity in the ensemble, which is far from being our case with only 5 identical models. Therefore, we have not implemented the separation of uncertainties for DE. In the case of HNN, the epistemic uncertainty is not modeled, so the separation of predictive uncertainty into aleatoric and epistemic makes no sense.

\section{Results and discussions} \label{sec:results}
\subsection{Summary metrics}
The performance of the methods in Section \ref{sec:methods} is evaluated against the 21 units of the N-CMAPSS test set with the metrics specified in Section \ref{sec:metrics}, and the hyperparameters listed in Table \ref{tab:bnn_params} and Table \ref{tab:nn_params}. The results aggregated over the runs are summarized in Table \ref{tab:res_method}.

\begin{table}
    \centering
    \begin{tabular}{lrrrrr}
        \hline
         & MAE $\downarrow$ & RMSE $\downarrow$ & NLL $\downarrow$ & RMSCE $\downarrow$ & Sharp $\downarrow$ \\
        \hline
DE & 6.277 $\pm$ 0.168 & \textbf{8.163 $\pm$ 0.136} & \textbf{2.372 $\pm$ 0.017} & 0.063 $\pm$ 0.018 & 10.676 $\pm$ 0.961 \\
MCD & 6.376 $\pm$ 0.143 & 8.466 $\pm$ 0.216 & 2.396 $\pm$ 0.024 & 0.071 $\pm$ 0.005 & 11.278 $\pm$ 0.424 \\
HNN & 6.203 $\pm$ 0.163 & 8.279 $\pm$ 0.181 & 2.516 $\pm$ 0.057 & \textbf{0.049 $\pm$ 0.015} & 8.733 $\pm$ 0.320 \\
RAD & 6.483 $\pm$ 0.102 & 8.619 $\pm$ 0.100 & 3.181 $\pm$ 0.533 & 0.051 $\pm$ 0.016 & 9.472 $\pm$ 0.494 \\
LRT & 6.403 $\pm$ 0.108 & 8.551 $\pm$ 0.122 & 4.553 $\pm$ 2.008 & 0.076 $\pm$ 0.017 & 9.031 $\pm$ 0.350 \\
FO & \textbf{6.158 $\pm$ 0.109} & 8.235 $\pm$ 0.113 & 5.007 $\pm$ 0.465 & 0.089 $\pm$ 0.012 & \textbf{8.581 $\pm$ 0.209} \\
        \hline
    \end{tabular}
    \caption{Performance metrics on the test set for all the methods described in Section \ref{sec:methods}. The units of RMSE and sharpness are in flight cycles. NLL and RMSCE are unitless. Rows are sorted by increasing NLL, and smaller values are better ($\downarrow$).}
    \label{tab:res_method}
\end{table}

In terms of NLL, which is the only metric accounting for both predictive accuracy and quality of uncertainty estimates, the performance of BNN models is worse than the one of frequentist alternatives. DE and MCD obtain the best NLL scores, but their largest sharpness can be a drawback for decision-making. In the case of BNN models, NLL variance across runs is orders of magnitude higher than in other methods, especially for the FO method. We can observe good NLL scores have a good correlation with the most conservative estimates of uncertainty, which is in line with the findings in \cite{ovadia2019can, quinonero2005evaluating, tran2020methods}. Therefore, if we rank models on NLL alone, we will clearly favor those yielding under-confident rather than over-confident uncertainty estimates. 

FO is the method with the best sharpness but at the cost of the worst calibration leading to the poorest RMSCE and NLL. On the contrary, HNN and RAD calibration are the best. In spite of its relative simplicity, HNN outperforms BNNs in terms of NLL, and offers the best calibration along with a competitive sharpness. As for accuracy, no method clearly outperforms the others when looking at the metrics in Table \ref{tab:res_method}, and differences in the scores are sometimes close and statistically insignificant. For instance, FO, HNN, and DE have got the top accuracy, but the other methods fall close behind in terms of MAE and RMSE. 

\subsection{Metrics per dataset}

The performance per dataset averaged over the results of all methods is summarized in Table \ref{tab:res_ds}.
Although aggregated performance per dataset is not useful for method ranking, it can help focus method performance analysis on the most challenging data subsets.   
In that sense, \texttt{D3} deserves special attention as it has the highest NLL score. Also, a large NLL variance points to substantial differences depending on the model. On the other hand, \texttt{D4} has the biggest MAE and RMSE scores, along with a large uncertainty (sharpness) in the predictions. However, a relatively moderate NLL and the lowest RMSCE indicates a good calibration of its predictive uncertainty. Finally, \texttt{D1} and \texttt{D5} seem overall less problematic for the models, whereas model predictions for \texttt{D2} together with \texttt{D3} are the worst calibrated.

\begin{table}[]
    \centering
    \begin{tabular}{lrrrrrrrrrr}
        \hline
         & MAE $\downarrow$ & RMSE $\downarrow$ & NLL $\downarrow$ & RMSCE $\downarrow$ & Sharp $\downarrow$ \\
        \hline
D2 & 6.129 $\pm$ 0.470 & 8.193 $\pm$ 0.668 & 2.848 $\pm$ 1.121 & \textbf{0.138 $\pm$ 0.057} & 7.323 $\pm$ 1.032 \\
D1 & 4.935 $\pm$ 0.281 & 6.522 $\pm$ 0.305 & 2.856 $\pm$ 1.198 & 0.061 $\pm$ 0.039 & 8.812 $\pm$ 1.066 \\
D4 & \textbf{9.126 $\pm$ 0.683} & \textbf{11.366 $\pm$ 0.597} & 2.944 $\pm$ 0.169 & 0.034 $\pm$ 0.018 & \textbf{12.893 $\pm$ 1.382} \\
D5 & 4.354 $\pm$ 0.312 & 5.769 $\pm$ 0.310 & 3.013 $\pm$ 1.058 & 0.076 $\pm$ 0.061 & 8.862 $\pm$ 1.279 \\
D3 & 6.063 $\pm$ 0.208 & 7.815 $\pm$ 0.343 & \textbf{4.32 $\pm$ 3.008} & 0.132 $\pm$ 0.065 & 7.729 $\pm$ 1.046 \\
        \hline
    \end{tabular}
    \caption{Performance metrics aggregated per dataset in the test set. Rows are sorted by increasing NLL and smaller values are better ($\downarrow$). The worst scores are highlighted in bold.}
    \label{tab:res_ds}
\end{table}

\begin{figure}[b!]
    \centering
    \includegraphics[width=0.65\textwidth]{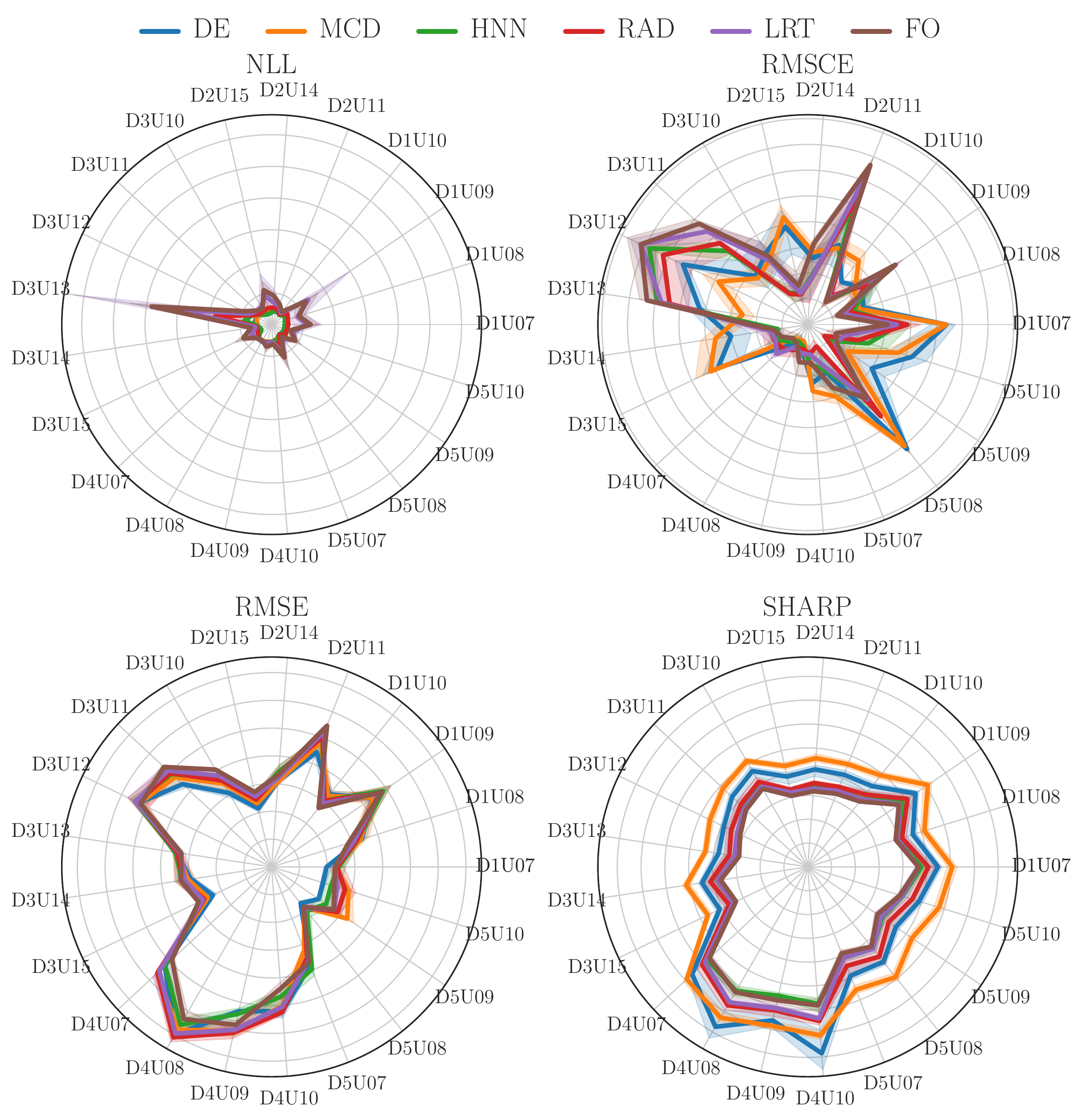}
    \caption{Performance metrics aggregated per engine unit and method.}
    \label{fig:unit_method_metrics}
\end{figure}

Figure \ref{fig:unit_method_metrics} provides an overview of method metrics per unit in the test set. The goal is to spot where method performance diverges the most and which units are the most challenging for each method. For instance, we can see that if \texttt{D3} has the worst NLL score, this is greatly due to the single contribution of \texttt{D3U13} and the BNN methods, especially LRT and FO. We can also observe FO is causing the large NLL variance. The significantly higher NLL score of this unit is consistent with its poor calibration (RMSCE) and indicates its sharpness (the lowest of all units) seems overly optimistic for the BNN. The overconfidence of these models with \texttt{D3U13} could be explained by the confusing presence in the training set of a unit whose degradation may look similar, but it is actually different, for instance in terms of number of cycles (see Figure \ref{fig:theta_vs_cycle}). 

On the contrary, RMSE scores are quite homogeneous across the methods. Also, all agree that unit \texttt{D4U08} is one of the hardest to predict accurately, whereas unit \texttt{D5U08} is one of the easiest ones. In terms of accuracy, units \texttt{D4U08}, \texttt{D4U09}, \texttt{D4U07}, \texttt{D3U12} and \texttt{D2U11} present the highest RMSE scores.

Concerning the calibration, we can observe significant differences across the methods depending on the unit. For instance, RMSCE for the BNN and HNN models is relatively higher compared to the other methods in units \texttt{D2U11}, \texttt{D3U11} and \texttt{D3U12}. However, for DE and MCD, \texttt{D2U15}, \texttt{D3U14} and \texttt{D3U15} calibration is much worse. As for the sharpness, method performance per unit is similar, with DE and MCD yielding systematically the most conservative scores.   

\subsection{Metrics per flight class}

As stated in \cite{arias2021aircraft}, each engine unit is assumed to operate only one of the three flight classes, i.e. short (\texttt{FC1}), medium (\texttt{FC2}) and long-haul (\texttt{FC3}) flights. Table \ref{tab:datasets} shows the distribution of flight classes in the training, validation and test subsets. The distribution in the training set is highly unbalanced in favour of \texttt{FC3} which represents 62\% of the samples, followed by \texttt{FC2} (27\%) and \texttt{FC1} (11\%). The validation set distribution is very similar to the training set. The test set is a bit more balanced with \texttt{FC3} representing 49\%, \texttt{FC2} 34\% and \texttt{FC1} 17\%. 

\begin{figure}
    \centering
    \includegraphics[width=0.8\textwidth]{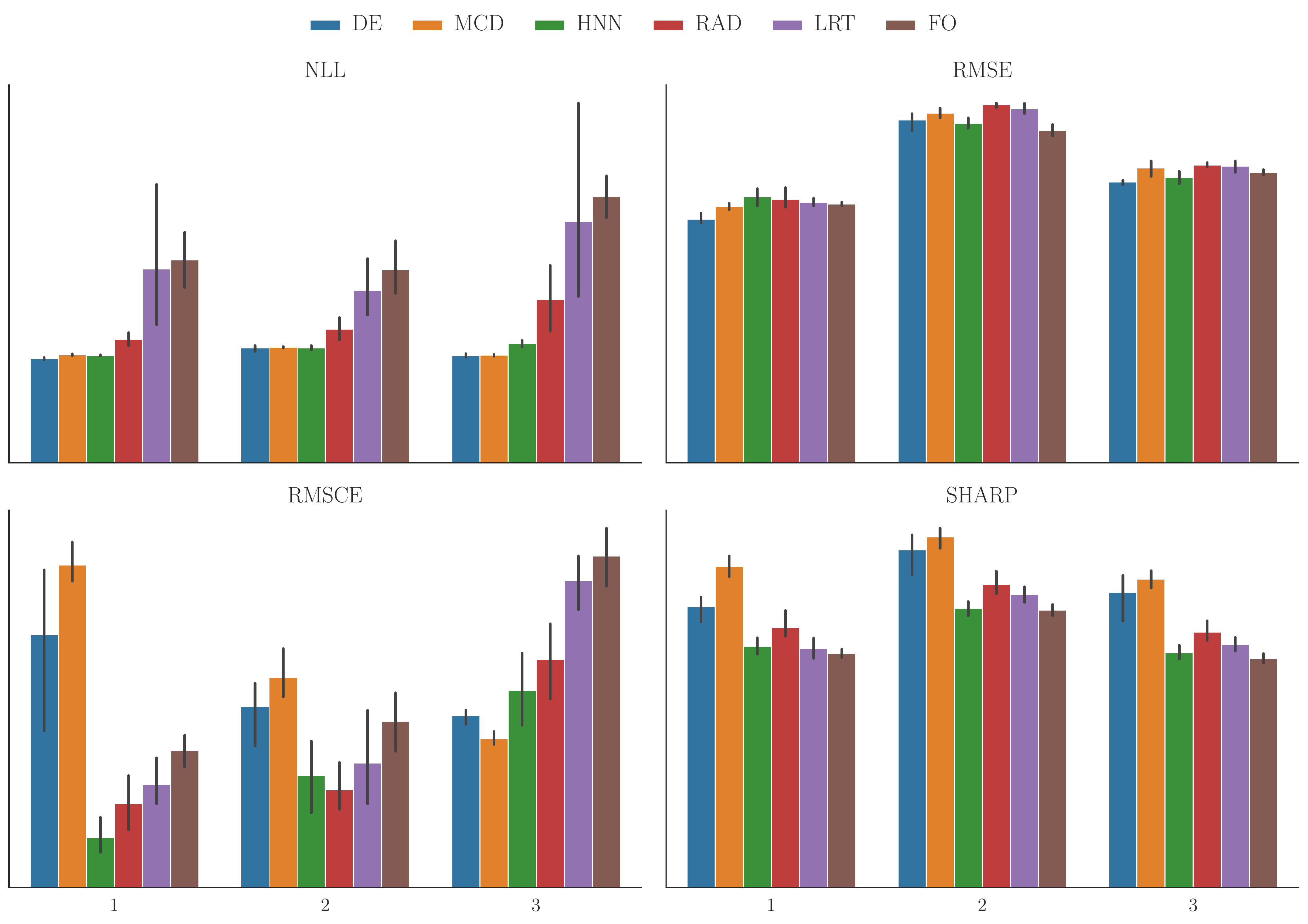}
    \caption{Performance metrics aggregated per flight class and method.}
    \label{fig:fc_method_metrics}
\end{figure}

Therefore, we can expect this unbalanced training set to have a negative impact in the performance of \texttt{FC1} units. Yet, we do not observe that when looking at Figure \ref{fig:fc_method_metrics}, which shows the method performance metrics across flight classes. In terms of accuracy, \texttt{FC2} followed by \texttt{FC3} are actually the classes with the worst scores. However, this is highly influenced by the contribution of \texttt{D4}, which is specific in terms of failure mode and has no \texttt{FC1}. When we removed \texttt{D4}, \texttt{FC1} became one with the largest sharpness and the second-worst accuracy.

Other than that, the important fact to note is the difference in calibration performance  across flight classes between DE and MCD and the BNN techniques. DE and MCD have a much worse calibration with \texttt{FC1}, whereas BNNs present a significantly higher RMSCE with \texttt{FC3}. However, the NLL plot tells a very different story, with DE and MCD yielding systematically better scores than the BNN models for all flight classes. The relative conservative sharpness and slightly better accuracy of DE and MCD translate into a surprisingly better NLL score than expected. This is an example on how the NLL score can be quite misleading by favoring the most conservative models.

\subsection{Quality of predictive uncertainty}
\begin{figure}
    \centering
    \includegraphics[width=0.48\textwidth]{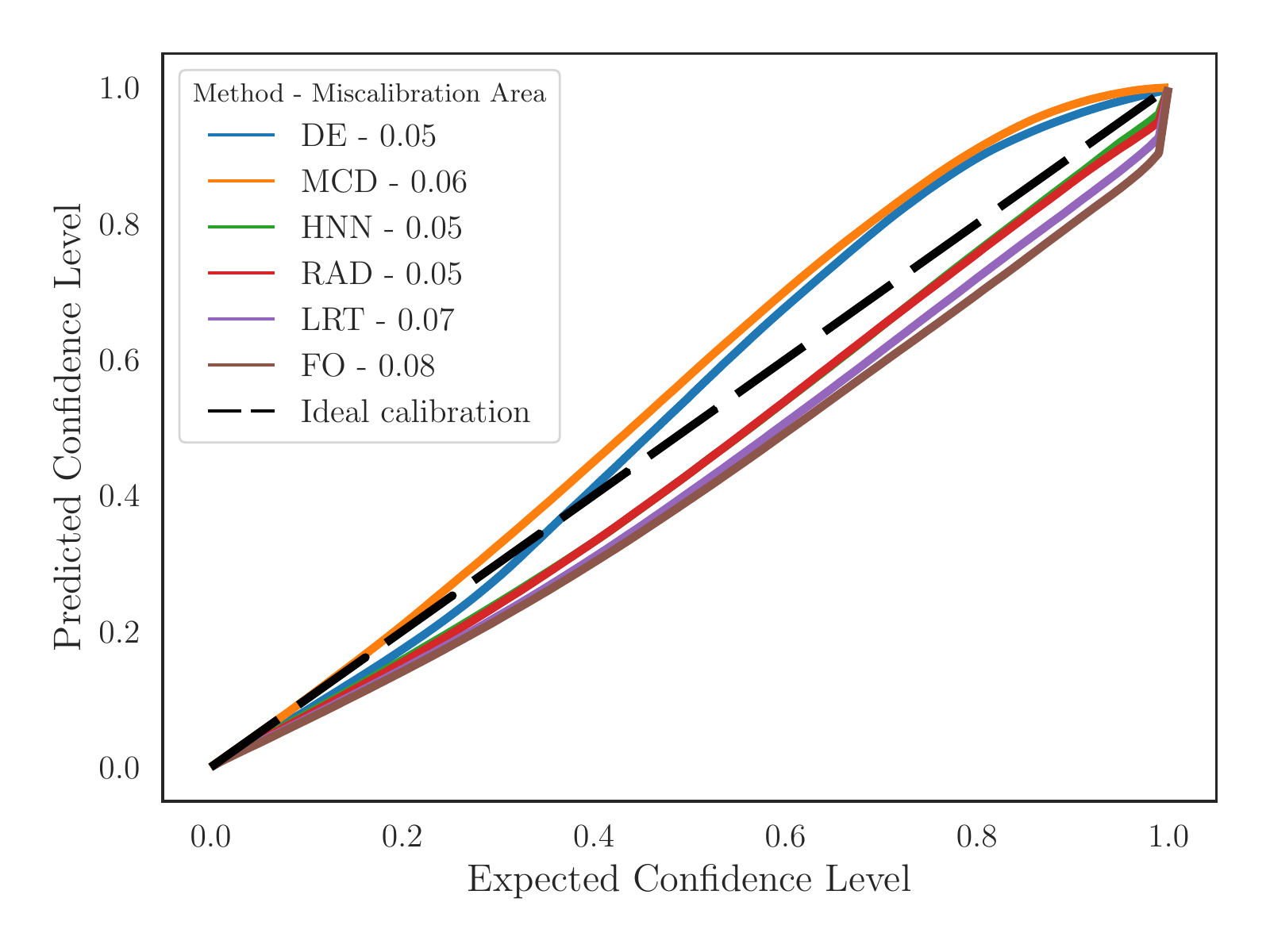}
    \includegraphics[width=0.48\textwidth]{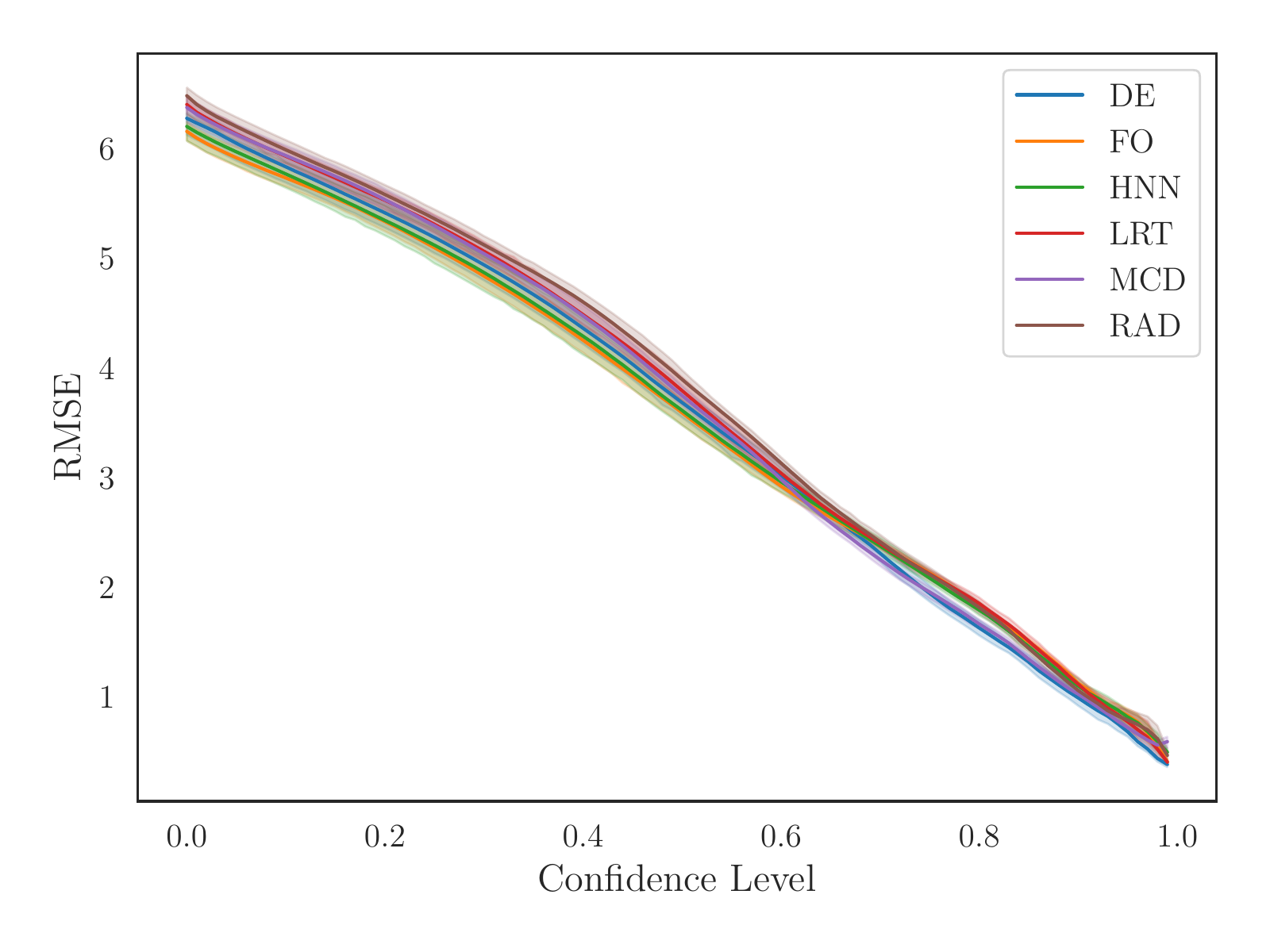}
    \caption{(Left) Method uncertainty calibration curves (ideal: $y=x$) and miscalibration area. (Right) RMSE versus confidence. Higher confidence levels correspond with lower predictive uncertainty ($\sigma^2$). The desirable inverse trend between RMSE and confidence level is observed across all methods.}
    \label{fig:cal_rmse_confidence}
\end{figure}

The calibration curves and the miscalibration area per method on the test units are displayed in the left plot in Figure \ref{fig:cal_rmse_confidence}. LRT and FO are the most overconfident (confident intervals too narrow). On the contrary, MCD and DE are under-confident (confident intervals too wide), especially from the mid-end of the system lifetime. FO has the highest miscalibration area, whereas RAD and HNN has the lowest ones, which is close to the ranking obtained with the RMSCE score.

For decision-making, a highly desirable property is that we can trust a model when the confidence in the prediction is high so that we can use a back-up solution when it is not (e.g., demand for human intervention). To evaluate this capability, we can plot model accuracy as a function of the confidence in the predictions, as shown in the right plot in Figure \ref{fig:cal_rmse_confidence}. This plot allows us to evaluate the model only when its confidence is above a certain threshold, or equivalently when the predictive uncertainty or variance ($\sigma^2$) is below a certain threshold. The higher the model confidence (the lower the predictive uncertainty), the better RMSE should be. Indeed, this inverse trend is exactly what we observe for all the methods, without any significant difference among them. 

\begin{figure}
    \centering
    \includegraphics[width=0.48\textwidth]{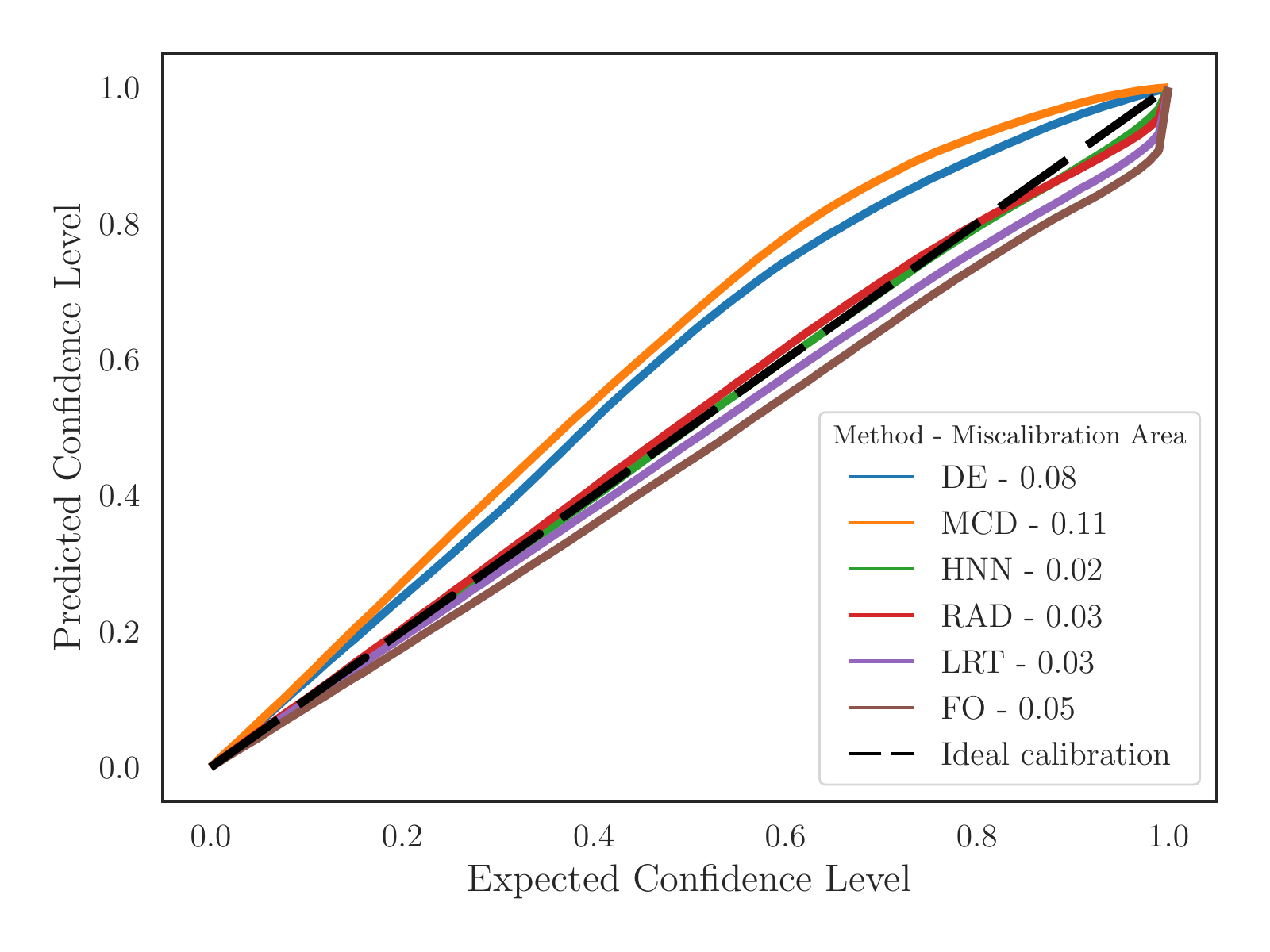}
    \includegraphics[width=0.48\textwidth]{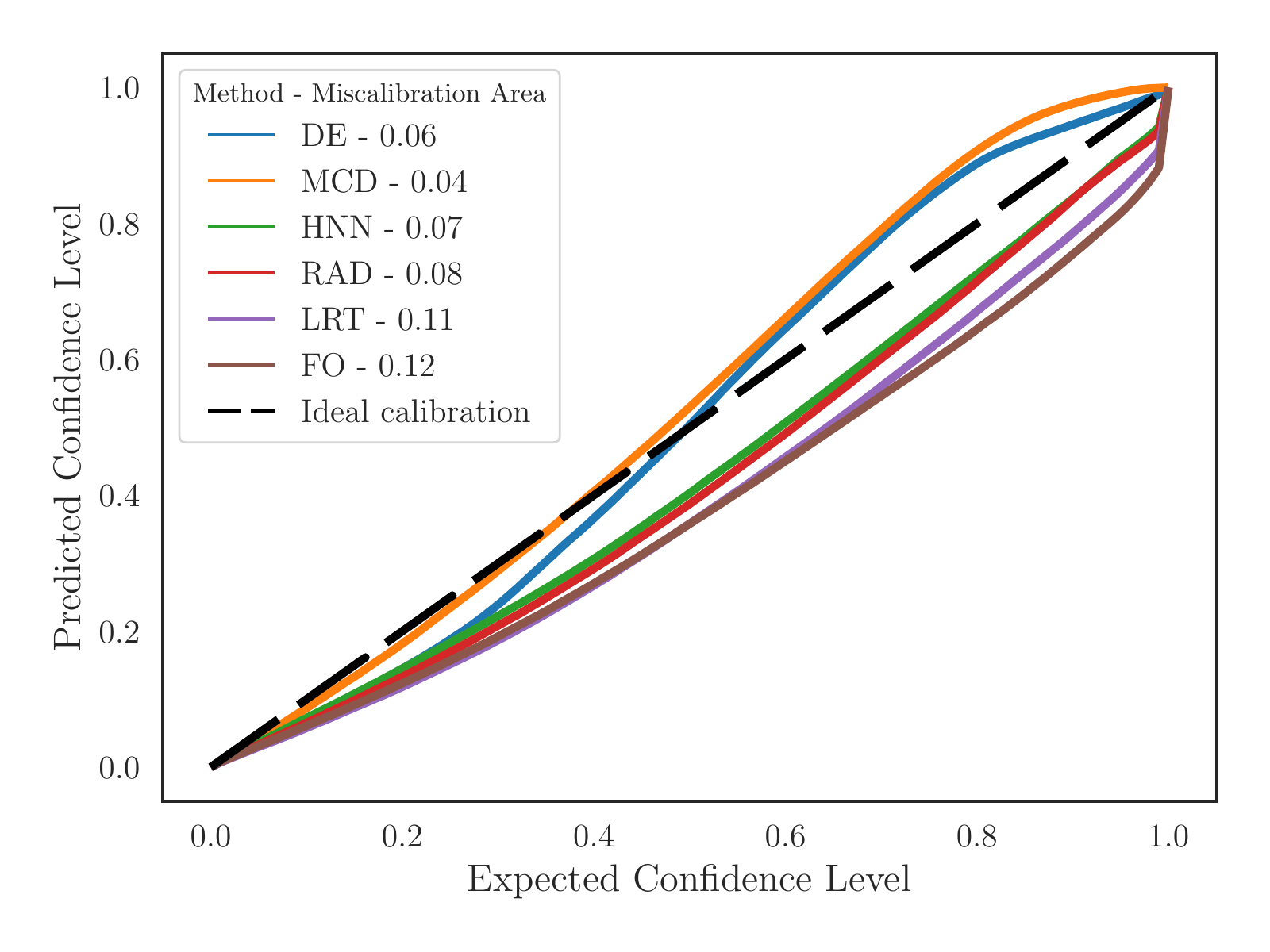}
    \caption{Calibration comparison between flight classes: \texttt{FC1} units (left) and \texttt{FC3} units (right).}
    \label{fig:calibration_fc}
\end{figure}

It is worth further exploring what happens with the calibration of flight classes \texttt{FC1} and \texttt{FC3}, given the observed differences in RMSCE between DE and MCD on one side and the BNNs on the other. Figure \ref{fig:calibration_fc} provides a comparison between the calibration of the two flight classes. HNN and BNN calibrations are excellent for \texttt{FC1} units, whereas DE and MCD suffer from under-confidence. For \texttt{FC3} units, HNN and BNNs are far overconfident, and DE and MCD calibrations are better than for \texttt{FC1} units, especially in the case of MCD.   

Another way of measuring \textit{if a model knows what it knows} concerns the generalization or robustness of the predictive uncertainty. This is usually evaluated in cases of domain shift between the train and the test set (OOD data). In OOD units, we expect high sharpness score or entropy~\footnote{With Gaussian likelihoods, sharpness, and entropy are related as both are a function of the predictive variance} combined perhaps with low accuracy. However, these conditions do not necessarily occur only in OOD situations, whose confirmation requires analyzing the differences in the operating parameters and degradation profiles between the train and test datasets (see Section \ref{sec:datasets}). 

\begin{figure}
    \centering
    \includegraphics[width=0.4\textwidth]{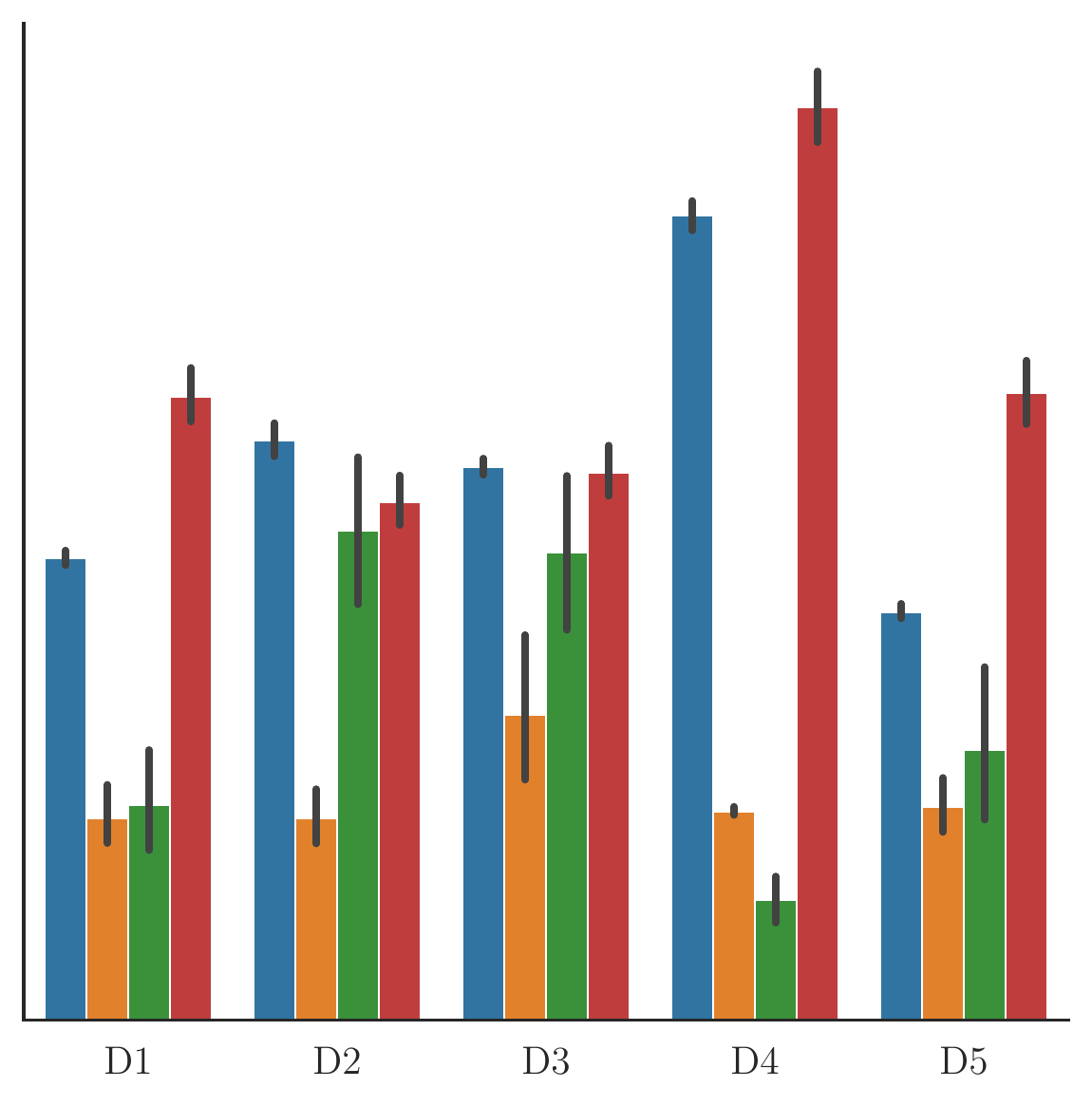}
    \includegraphics[width=0.44\textwidth]{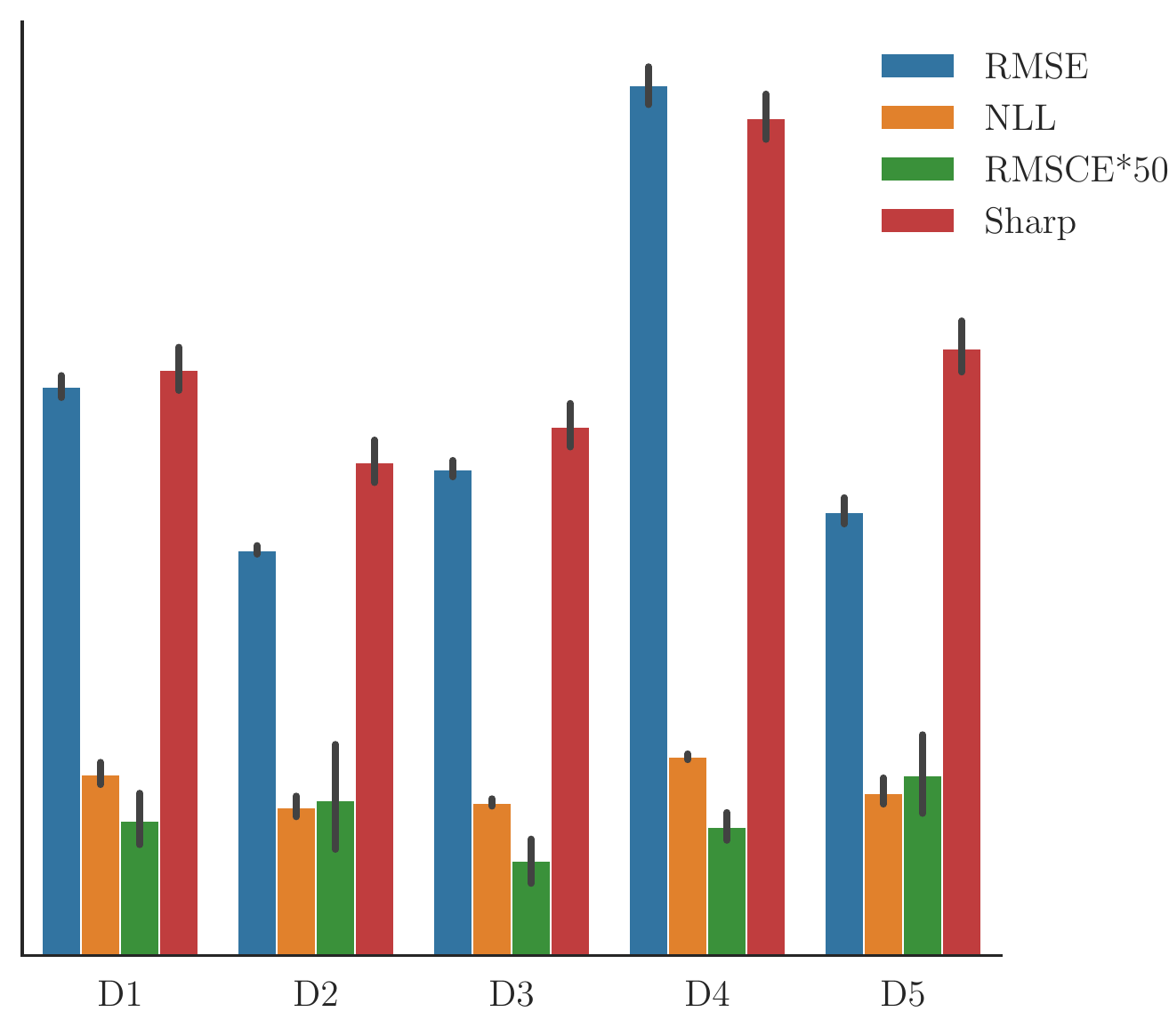}
    \caption{Metrics per dataset for the test (left) and training (right) sets. \texttt{D4} RMSE and sharpness is already the worst in training set, suggesting large sharpness in the test set may not be only linked to OOD.}
    \label{fig:ds4}
\end{figure}

For instance, we see in Figure \ref{fig:unit_method_metrics} that units with the largest sharpness are the ones in \texttt{D4}. We showed in Section \ref{sec:datasets} this is due to the fact that \texttt{D4} is a challenging dataset because of its specificities. The right plot in Figure \ref{fig:ds4} shows \texttt{D4} has also the worst accuracy and sharpness in the train set, which further supports the assumption that what happens with \texttt{D4} as a whole cannot be simply explained in terms of distributional shift. 

\begin{figure}[t!]
    \centering
    \includegraphics[width=0.48\textwidth]{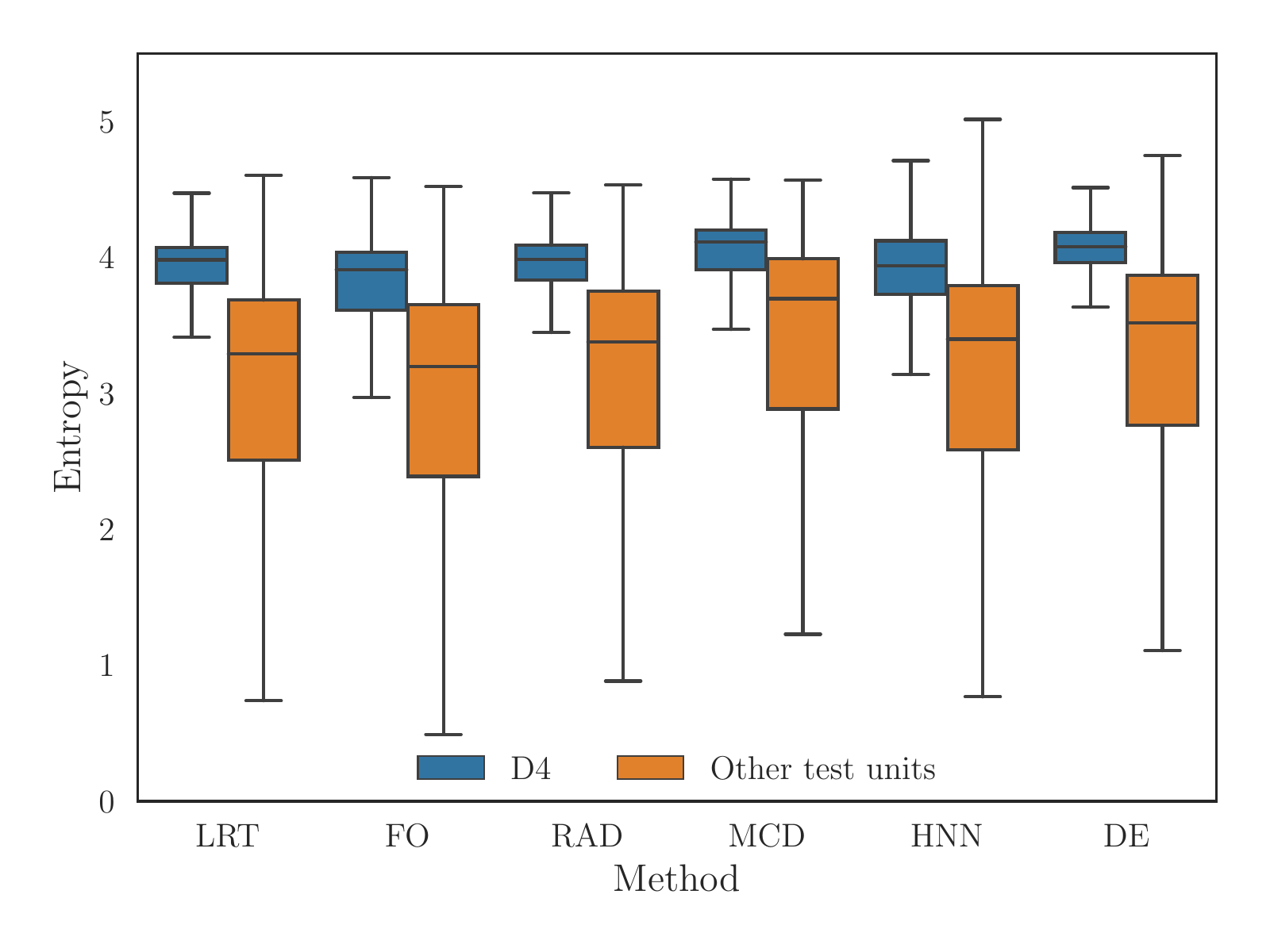}
    \includegraphics[width=0.48\textwidth]{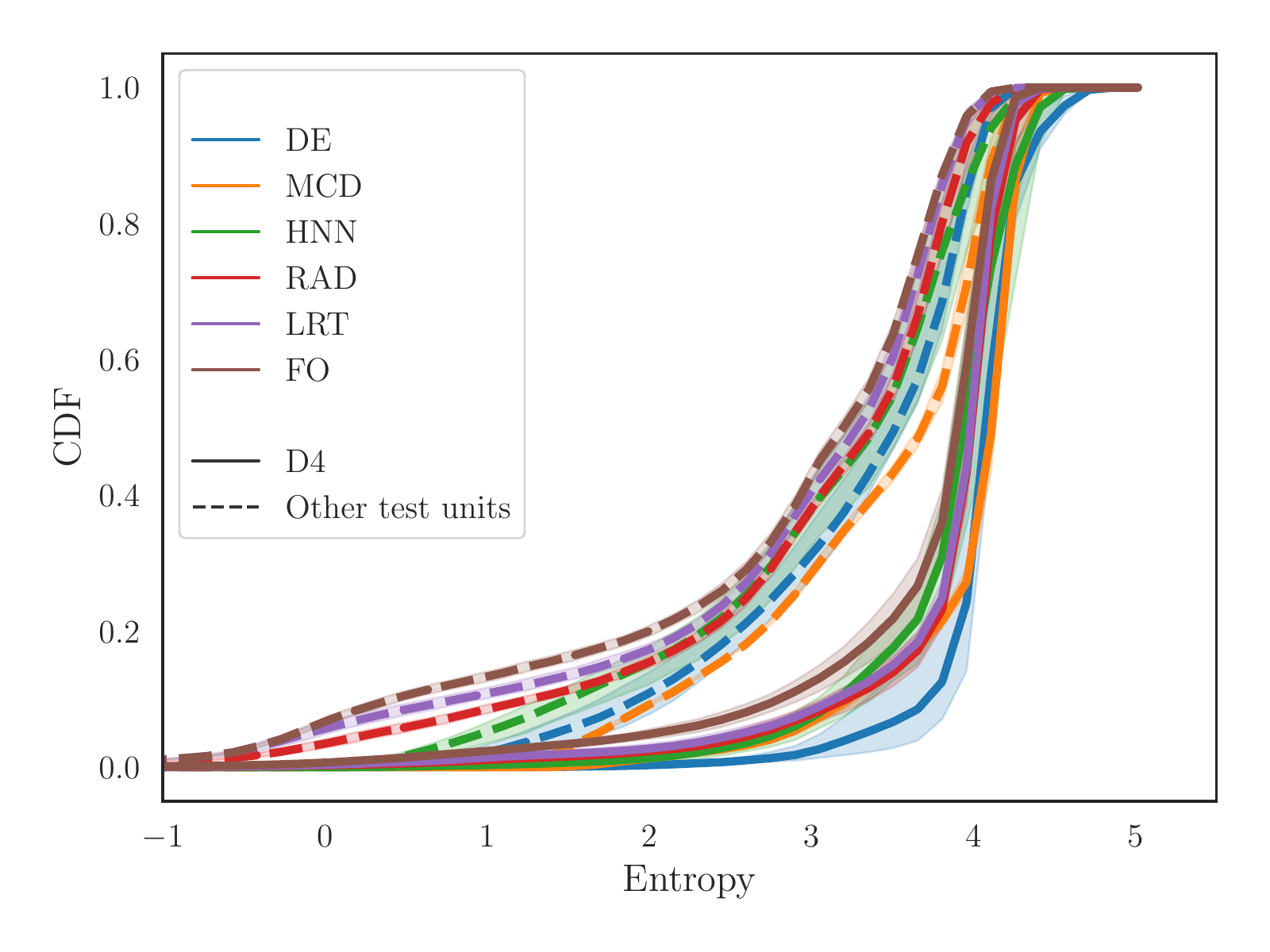}
    \caption{Entropy comparison between \texttt{D4} test units and the rest of test units. All methods estimate higher uncertainty levels for \texttt{D4} units.}
    \label{fig:entropy_d4}
\end{figure}

Indeed, this illustrates rather a case of robustness of predictive uncertainty to a dataset where degradation is genuinely harder to predict. All methods are robust in the sense their high RMSE is linked to large sharpness and good calibration (low RMSCE). We can better visualize what happens by comparing the entropy distributions of the \texttt{D4} test units vs the rest of test units. We see in Figure \ref{fig:entropy_d4} all methods exhibit higher entropy levels for the \texttt{D4} units. The degree of separation in the entropy distributions depends on the method, i.e., it is less evident with MCD and HNN than with RAD, LRT and DE.

\begin{figure}[t!]
    \centering
    \includegraphics[width=0.48\textwidth]{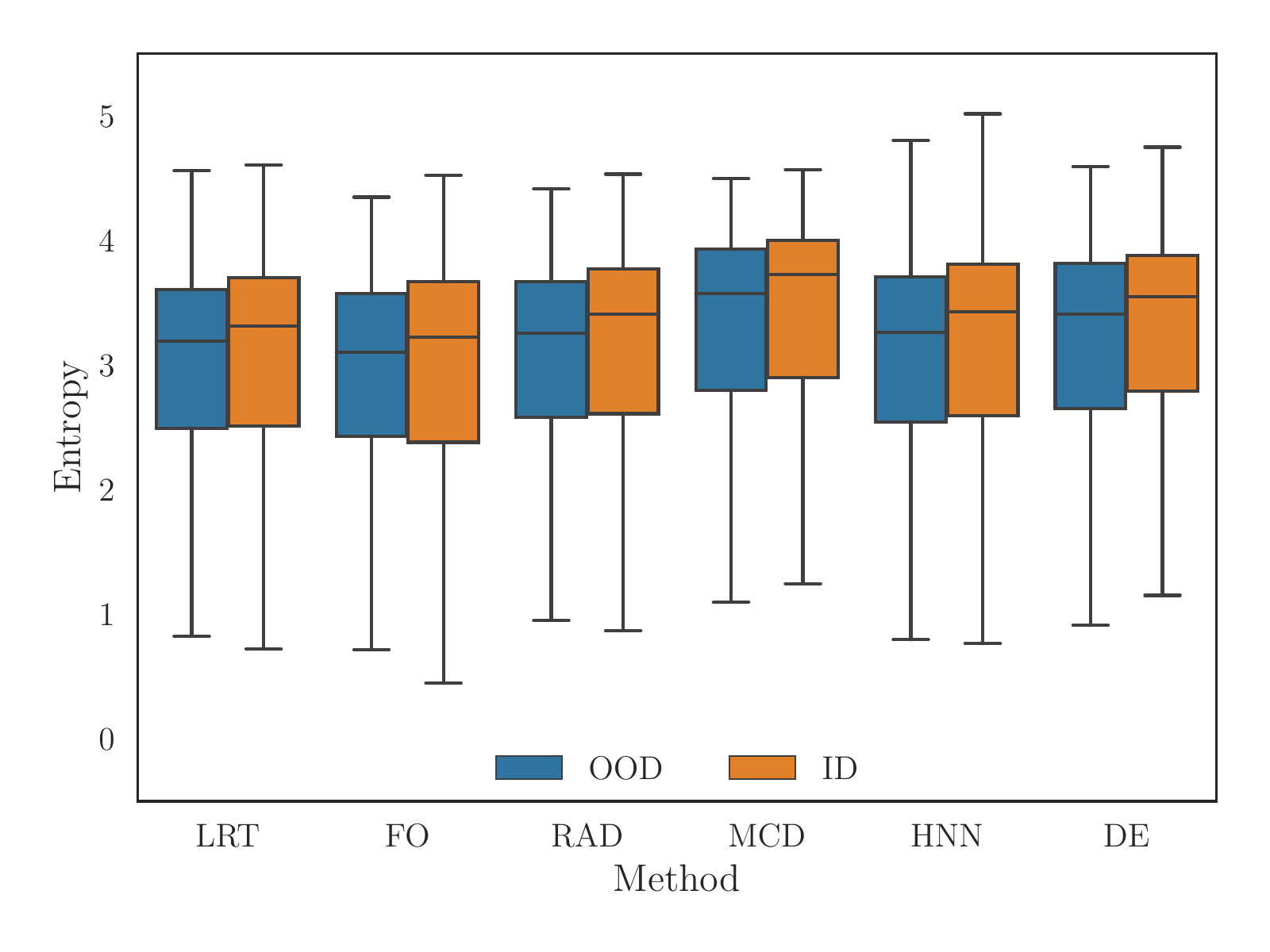}
    \includegraphics[width=0.48\textwidth]{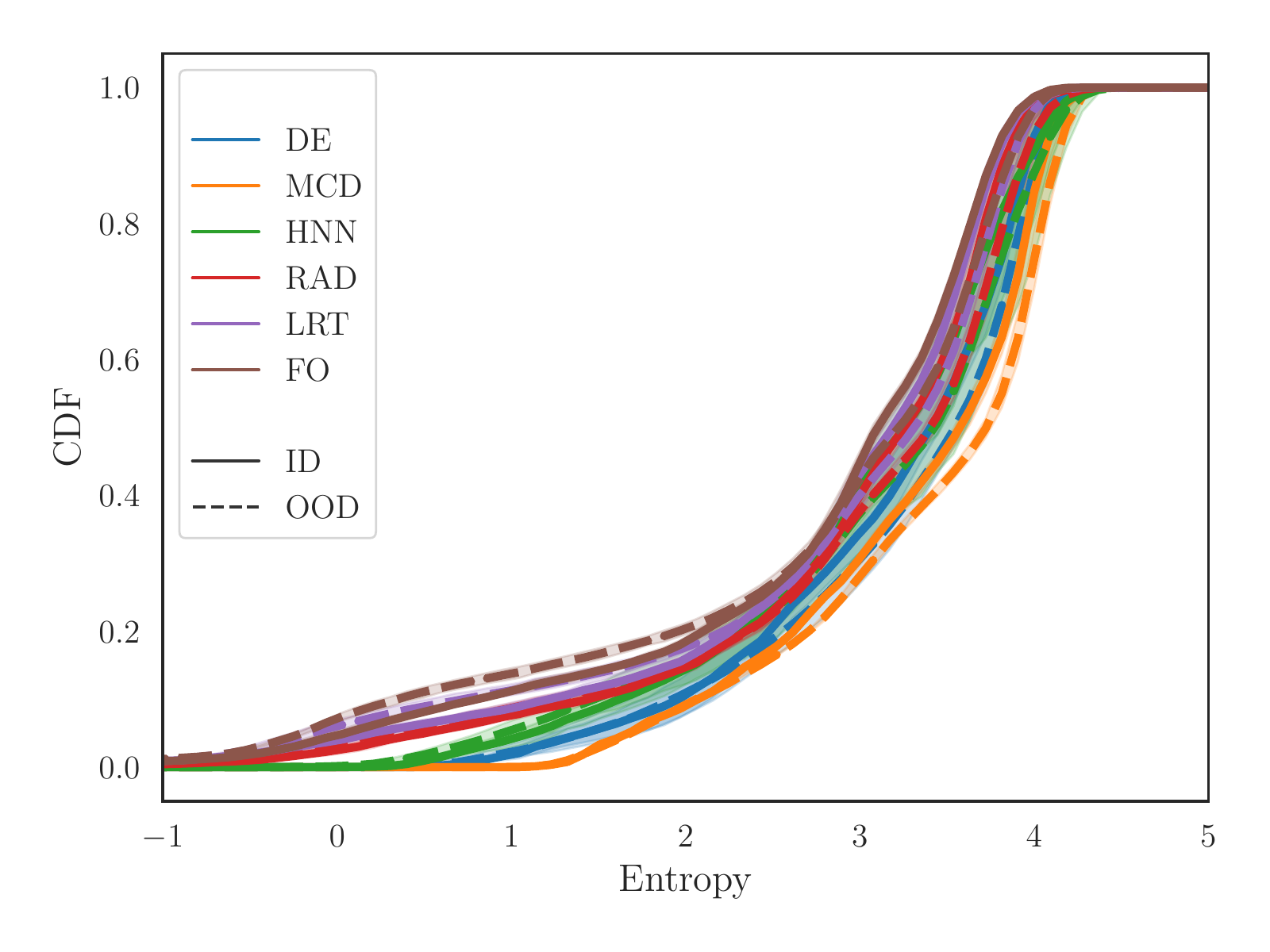}
    \caption{Entropy comparison between in-distribution (ID) and out-of-distribution(OOD) test units (\texttt{D2U11}, \texttt{D3U11} and \texttt{D3U12}). All methods estimate lower levels of entropy for OOD units due to bad calibration (overconfidence).}
    \label{fig:entropy_ood}
\end{figure}

In Section \ref{sec:datasets}, we identified \texttt{D2U11}, \texttt{D3U11} and \texttt{D3U12} as potential OOD cases because of their degradation profiles being significantly different from the rest. However, Figure \ref{fig:unit_method_metrics} shows that sharpness is not as high as it should be for these OOD units. What is actually high instead is both their RMSE and RMSCE, the latter pointing to calibrations being among the worst ones, especially for the BNNs and HNN models. Figure \ref{fig:entropy_ood} allows us to compare the entropy distributions of ID and OOD units. Unfortunately, all methods exhibit indeed an entropy which is slightly lower for the OOD units, which is due to overconfidence. Thus, robustness of the methods to OOD data is severely compromised by bad calibration associated with poor accuracy. We think these results are significant and should be further investigated in future work.

\subsection{Method performance over the system lifetime}

\begin{figure}
    \centering
    \includegraphics[scale=0.5]{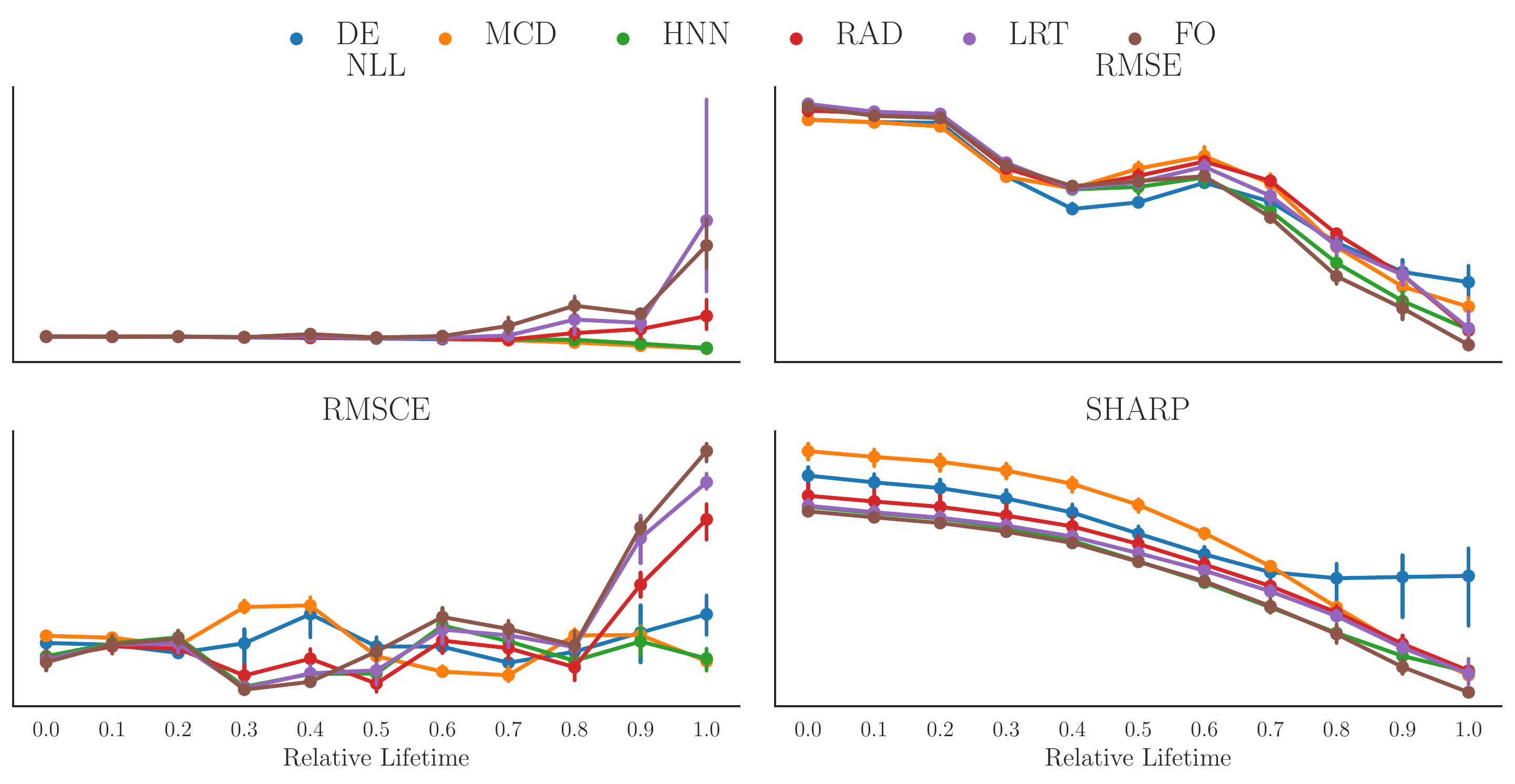}
    \caption{Method performance evolution over the system lifetime.}
    \label{fig:rlt_metrics}
\end{figure}

Reliability of RUL predictions is obviously most needed when a system reaches the end of life, where wrong predictions have a higher economic and safety cost. Figure \ref{fig:rlt_metrics} shows the evolution of the metrics aggregated over all the test units. The downward trend in RMSE and sharpness achieved by all methods is reassuring in the sense predictions become more accurate and certain by the end, especially in the case of HNN and BNNs. However, we can observe a change from a downtrend to an uptrend in the DE sharpness at around 70\% of the system lifetime. More worrying is the surge at the end in the RMSCE and NLL scores of the BNN models.     

\begin{figure}
    \centering
    \includegraphics[scale=0.23]{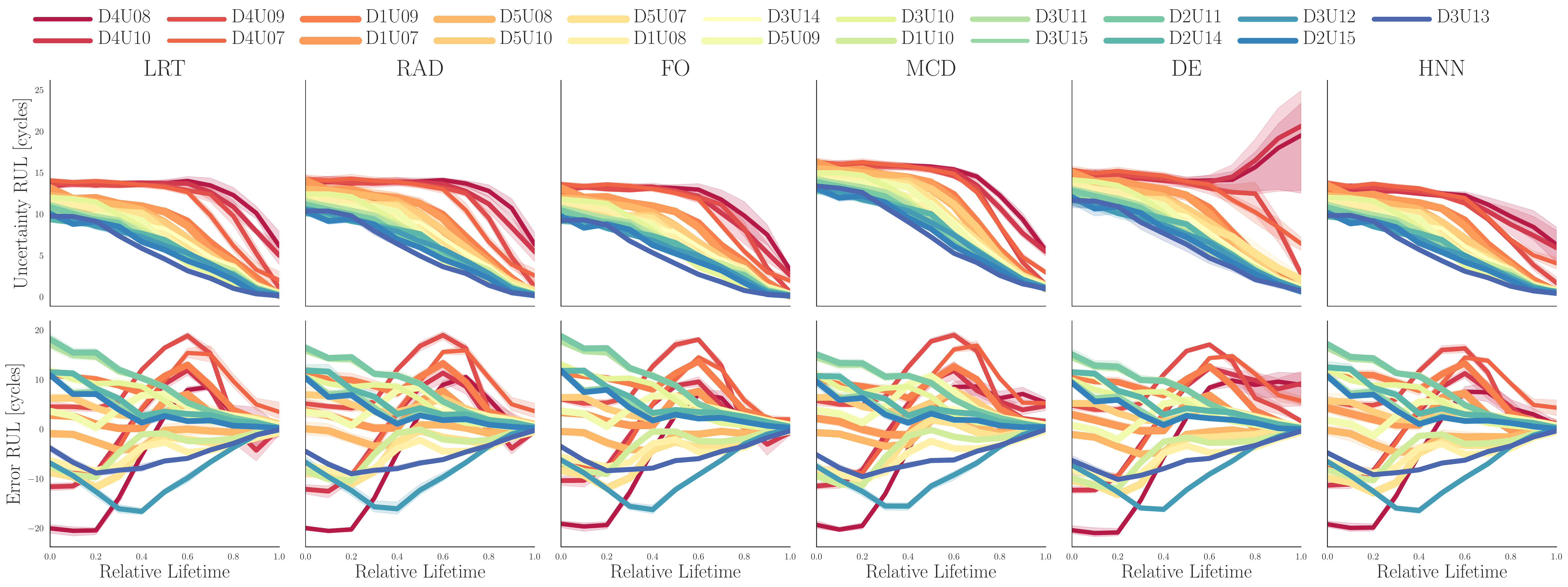}
    \caption{Evolution of predictive uncertainty and prediction error per method on the test set units.}
    \label{fig:unit_method_std_err}
\end{figure}

The evolution of predictive uncertainty and prediction error (see Figure \ref{fig:unit_method_std_err}) follows consequently a downward trend as well for most of the units. Again, we see how the uncertainty and predictive errors are higher for \texttt{D4} units, especially for \texttt{D4U08} and \texttt{D4U09}. For these units, the pattern of evolution differs, and a peak in error can be observed in all methods at around 60\% of the system lifetime. In the case of DE, the surge on the uncertainty at the later cycles towards the end of life is to be mostly attributed to \texttt{D4U08} and \texttt{D4U09} units.

\begin{figure}
    \centering
    \includegraphics[scale=0.38]{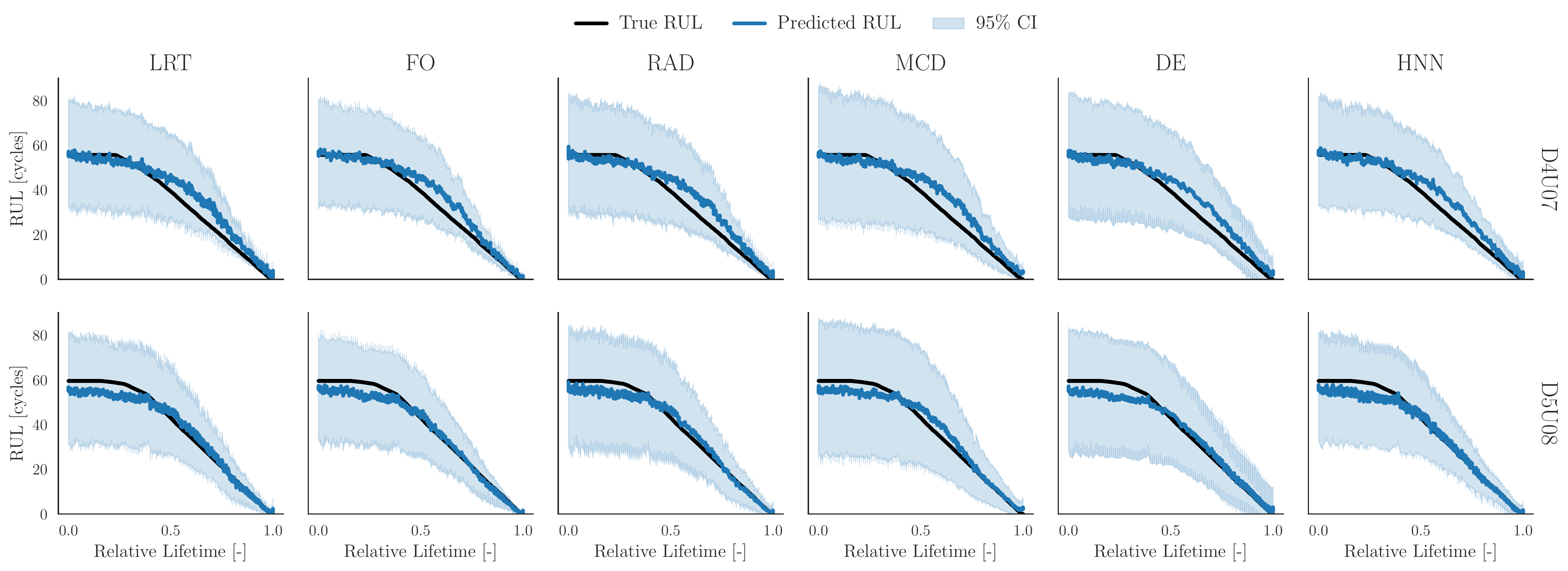}
    \caption{RUL estimations for units \texttt{D4U07} and \texttt{D5U08}.}
    \label{fig:rul}
\end{figure}

As an illustrative example, Figure \ref{fig:rul} compares the RUL predictions for one of the difficult units (\texttt{D4U07}) against one of the easiest ones (\texttt{D5U08}). The blue line (predicted RUL) follows the black line (true RUL) more or less accurately, depending on the unit and the method. In all cases, predictive uncertainty is reduced when approaching the end of life, as it is shown with the 95\% confidence interval in light blue. 

\subsection{Epistemic and aleatoric UQ}
\begin{figure}
    \centering
    \includegraphics[scale=0.5]{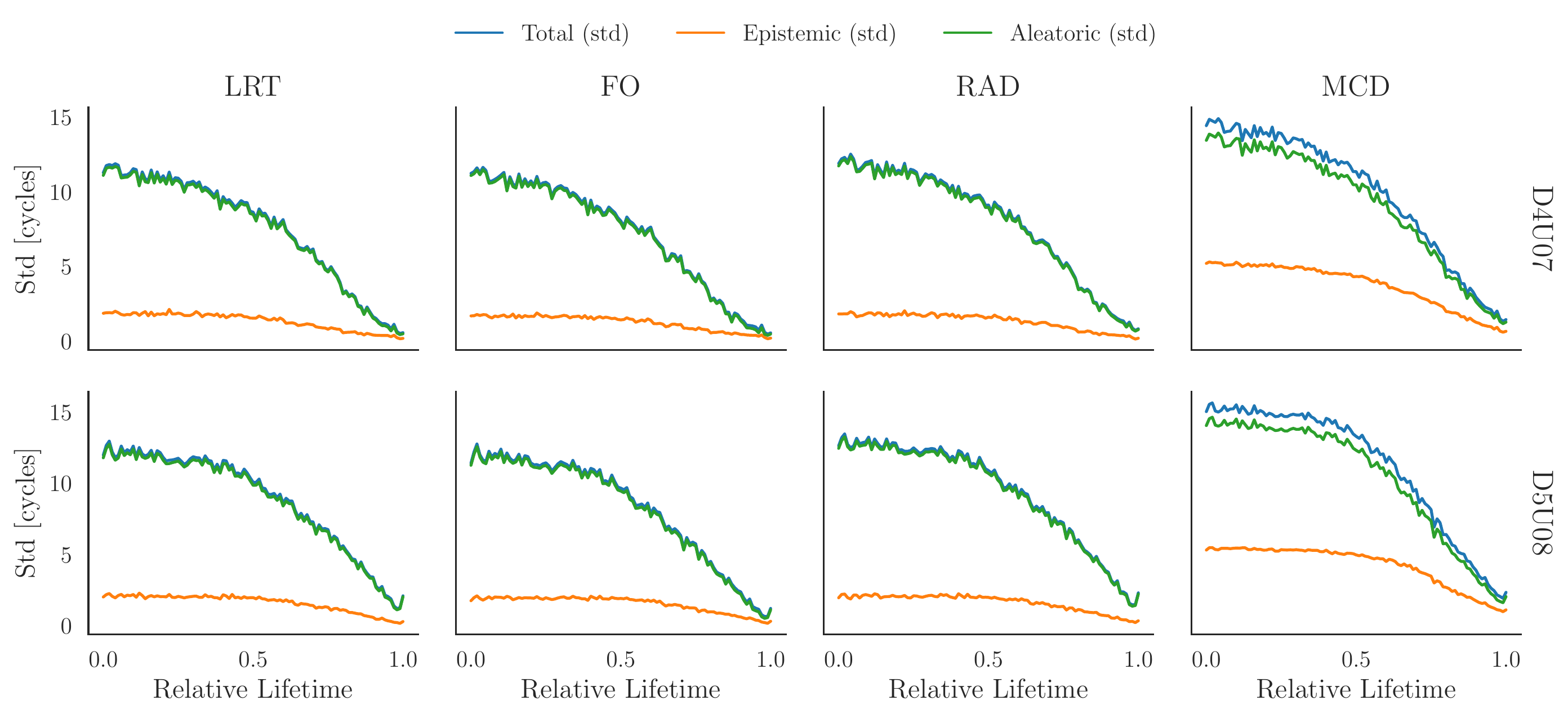}
    \caption{Break-down of predictive uncertainty into epistemic and aleatoric uncertainty for units \texttt{D4U07} and \texttt{D5U08}.}
    \label{fig:eps_al_uq}
\end{figure}

Figure \ref{fig:eps_al_uq} provides a comparison of the decomposition of the predictive uncertainty into aleatoric and epistemic uncertainty as per Equation \ref{equ:ep_al_unc} for \texttt{D4U07} and \texttt{D5U08}. As we are using standard deviations instead of variances, the sum of the aleatoric (green) and epistemic (orange) uncertainties is not equal to the total uncertainty (blue). The dominance of the aleatoric uncertainty over the epistemic one is very clear, especially for the BNN. The resulting decomposition is slightly dependent on the method, but very similar for the three BNN techniques. Differences are minor in the decomposition between an easy or a difficult-to-predict unit. We see a clear decreasing trend in the aleatoric uncertainty over the time, whereas the epistemic one remains relatively more constant, which is in agreement with the results in \cite{li2020bayesian}. For aleatoric uncertainty, the explanation lies in the accumulative nature of noise over the lifetime of the system. This is not the case with the epistemic uncertainty, which only depends on the availability of training data covering the system's lifetime. 

\section{Conclusion} \label{sec:conclusion}
Following a literature review, we examined a set of UQ methods for deep learning RUL estimation, including recent variational BNN methods for gradient variance reduction as well as alternatives like MCD, DE and a frequentist HNN. We have evaluated the performance of these methods with the recent N-CMAPSS dataset in terms of prediction accuracy and quality of the predictive uncertainty. All the considered methods scale relatively well with the size of N-CMAPSS and provide sensible uncertainty estimates and accurate RUL predictions in most of the test cases. However, we have not found any of the methods to be particularly robust to OOD cases, but BNN and HNN methods seem to suffer even more in OOD from low accuracy associated with overconfidence.  

Comprehensive evaluation of the methods based on a set of UQ metrics is not straightforward. There exists a trade-off between them, e.g., between calibration and sharpness. Even scoring rules like NLL, which consider both calibration and sharpness altogether, seems biased towards under-confident models. In other words, it is not possible to rank the methods based on a single metric, the existence of which is an open research problem. The evaluation is also challenging because method performance depends on the subset of data considered (e.g., unit, flight class).  

In spite of their complexity, we have not found BNN models to clearly outperform the simpler alternatives. Indeed, results show that HNN models can be a competitive and balanced alternative. HNN models are lighter, as well as much faster and more straightforward to train than BNN models. DE and MCD suffer from conservative uncertainty estimates, which may not be informative enough for decision-making. If we need to break down epistemic and aleatoric uncertainties (e.g. for active learning), MCD and RAD are probably the best choices to consider.

Based on the relatively good performance achieved by HNN, it would be interesting to include in our benchmark other non-Bayesian techniques to improve the vanilla HNN we used. For instance, deep evidential regression \cite{amini2020deep} or the improvements suggested in \cite{seitzer2022pitfalls, skafte2019reliable} would be worth considering. Another aspect to evaluate we excluded from the benchmark is the influence of likelihood functions other than the basic Gaussian, such as skewed Gaussian, the Weibull, or Logistic ones.

In future work, we think the robustness of the methods to OOD situations should deserve a deeper assessment. We could, for instance, exclude a flight class from the training set or use combinations of training and test units with different degrees of distributional shifts. Safety aspects related to the reliability of the RUL estimates towards the end of the system lifetime should be further investigated as well. Finally, the benchmark could be extended to include deep Gaussian process models as well, since the work in \cite{biggio2021uncertainty} is based only on the \texttt{D2} subset of N-CMAPSS making comparisons with our results difficult. 

\bibliographystyle{unsrtnat}
\bibliography{biblio.bib}
\end{document}